\documentclass[11pt,a4paper]{article}
\usepackage[hyperref]{acl2020}
\usepackage{times}
\usepackage{latexsym}
\usepackage{multirow}
\usepackage{sidecap}
\usepackage{amsthm}
\usepackage{amsmath}
\usepackage{amssymb}
\usepackage{amsfonts}
\usepackage{verbatim} % for multi-line comments
\usepackage{url}
\usepackage{pifont}
\usepackage{eqparbox}
\usepackage{arydshln} % for dashed line in tables 
\usepackage{bigstrut}
\usepackage[linewidth=1pt]{mdframed}
\usepackage{framed}
\usepackage{float}
\usepackage{adjustbox}
\usepackage{booktabs}
\usepackage{url}
\usepackage[normalem]{ulem}
\usepackage{tabu}
\usepackage{subcaption}

\usepackage{comment}
\usepackage[colorinlistoftodos]{todonotes}

\usepackage{microtype}
\usepackage{xspace,soul}

\aclfinalcopy % Uncomment this line for the final submission
\definecolor{purple}{rgb}{0.5,0,1}
\definecolor{dcyan}{rgb}{0.2,0.6,0.5}
\definecolor{light-gray}{gray}{0.95} % used in table

\definecolor{darkgreen}{RGB}{0,140,0}
\definecolor{darkred}{RGB}{200,0,0}
\definecolor{lightgreen}{RGB}{189,252,192}
\definecolor{lightred}{RGB}{255,205,212}
\definecolor{lightyellow}{RGB}{255,240,160}
\definecolor{lightblue}{RGB}{195,221,255}
\definecolor{lightpurple}{RGB}{232,209,255}
\definecolor{lightorange}{RGB}{252,193,177}

\newcommand{\orangetext}[1]{\colorbox{lightorange}{#1}\xspace}

\newcommand{\ignore}[1]{}
\newcommand{\qn}[1]{\textrm{\small \color{violet} [QN: #1]}}
\newcommand{\bz}[1]{\textrm{\small \color{orange} [BZ: #1]}}
\newcommand{\bzch}[1]{\textrm{\color{black} #1}}
\newcommand{\qnch}[1]{\textrm{\color{violet} #1}}
\newcommand{\danielk}[1]{{\color{blue} \small [DK: #1]}}

\newcommand{\mctaco}{\textsc{McTaco}}
\newcommand{\modelname}{\textsc{TacoLM}}
\newcommand{\keywordCode}[1]{{\small \texttt{#1}}}
\newcommand{\keywordText}[1]{{\textsf{#1}}}
\newcommand{\prob}[1]{$\mathbb{P}\left(#1\right)$}

\newcommand{\event}[1]{\textit{{#1}}}
\newcommand{\idxevent}[3]{\event{e#1:\textrm{\color{#3}#2}}}

\newcounter{exctr}
\newcommand{\addexctr}{\refstepcounter{exctr}\theexctr}
\newcounter{eventCtr}
\newcommand{\addeventCtr}{\refstepcounter{eventCtr}\theeventCtr}

\title{
\vspace*{-0.5in}
{{\small \hfill ACL'20}\\
\vspace*{.25in}} 
Temporal Common Sense Acquisition with Minimal Supervision \\
}

\author{Ben Zhou,$^1$ Qiang Ning,$^2$
Daniel Khashabi,$^{2}$ Dan Roth$^1$ \\
{ \normalsize  $^1$University of Pennsylvania, \;  $^2$Allen Institute for AI} \\
{\tt \footnotesize \{xyzhou,danroth\}@cis.upenn.edu \; \{qiangn,danielk\}@allenai.org}
}

\date{} 

\begin{document}
\maketitle
\begin{abstract}
\ignore{
Understanding the temporal aspects of events in the text is crucial for reading comprehension. Recent works on contextualized language models are capable of understanding these aspects to some extent, however, the use of independent discrete tokens and the difficulty of understanding prepositions limit such In this work, we propose a method that generates temporal event representations. Specifically, we use rule-based patterns to automatically label temporal arguments in free form texts and propose a model that uses such data to jointly represent multiple temporal aspects, including duration, frequency, typical time, and others. We show that the proposed model produces useful distributed semantics about time, and such representations can be applied to achieve better performances in several related tasks, including event hierarchy and more. }

Temporal common sense (e.g., duration and frequency of events) is crucial for understanding natural language.
%stories, 
However, its acquisition is challenging, partly because such information is often not expressed explicitly in text, \bzch{and human annotation on such concepts is costly.}
% \bzch{deeming costly human annotations necessary.}
%, and costly human annotations are deemed necessary. 
% \dr{Note that I changed the order of the presentation below to emphasize the key contribution.}
This work proposes a novel sequence modeling approach that exploits explicit and implicit mentions of temporal common sense, extracted from a large corpus, to build 
\modelname,\footnote{\url{https://cogcomp.seas.upenn.edu/page/publication_view/904}} a \textbf{t}empor\textbf{a}l \textbf{co}mmon sense \textbf{l}anguage \textbf{m}odel.
% \sout{, a time-aware language model}
\ignore{
This work first designs syntactic rules to retrieve both explicit and implicit mentions about temporal common sense from a large corpus.
% , both explicit ones (e.g., how long some event lasts) and implicit ones (e.g., one event is longer than the other); 
We then propose a novel sequence modeling approach that exploits these signals jointly and contextually, and build 
\bzch{\modelname \footnote{\url{https://github.com/CogComp/TacoLM}} (\textbf{t}empo\textbf{a}l \textbf{co}mmon sense \textbf{l}anguage \textbf{m}odel.)
% \sout{, a time-aware language model}
}
}
% \danielk{rename it to Taco-LM?}
Our method is shown to give quality predictions of various dimensions of temporal common sense (on UDST and a newly collected dataset from RealNews). It also produces representations of events for relevant tasks such as duration comparison, parent-child relations, event coreference and temporal QA
%question answering
(on TimeBank, HiEVE and \mctaco{}) that are better than using the standard BERT.
Thus, it will be an important component of temporal NLP.
% \bz{temporal-driven more restricted than temporal-aware, as HiEVE strictly is not temporal-driven.} 
%tasks.

\ignore{
exploit multiple sources of cheap signals \qn{think of detailing it out} jointly from freely-available text, and we show that our approach not only gives quality prediction of various aspects of temporal common sense, but also produces better representation of events for relevant tasks than using language models only, such as parent-child relations between events and event coreference \qn{highlight some numbers}. We also discuss multiple details that are designed to fix flaws in existing models and how they benefit different tasks. We hope that this work will be a very useful component for a wide range of natural language applications.
}
\end{abstract}

\section{Introduction}
\label{sec:intro}
% \qn{FYI. In my communication with the ECE editorial office at UIUC, they suggested: common sense is a noun, while commonsense is an adjective. So it's ``temporal common sense'' and ``commonsense understanding/information.'' Take it with a grain of salt but be consistent ourselves.}
%time is important but heavily relies on common sense
Time is crucial when describing the evolving world. It is thus 
important
%an important task 
to understand time as expressed in natural language text.
Indeed, many natural language understanding (NLU) applications, 
%may 
including information retrieval, summarization, causal inference, and QA %question answering 
\cite{ULADVP13,CCMB14,LCUMAP15,BSCDPV16,LeeuwenbergMo17,NWPR18}, rely on understanding time.

% define: temporal common sense and example 1 (Porter) 
However, understanding time in natural language text heavily relies on \emph{common sense} inference.
Such inference is challenging since commonsense information is rarely made explicit in text (e.g., how long does it take to open a door?) Even when such information is mentioned, it is often affected by another type of
%due to the well-known 
{\em reporting bias}: 
%in natural language: 
people rarely say the obvious, in order to communicate more efficiently, but sometimes highlight rarities \cite{Schubert02,Durme09,GordenDu13,ZRDV17,BauerWaBa18,TDGYBC18}.
% \dr{note that I added the first part -- we cannot start without mentioning that TCS information isn't explicit and just jump to reporting bias (I view these as different).}

This is an even more pronounced phenomenon when it comes to temporal common sense (TCS) \cite{ZKNR19}. 
In Example~\ref{ex:tcs}, human readers know that a typical vacation is likely to last at least a few days, and they would choose \emph{``will not''} to fill in the blank for the first sentence; instead, with a slight change of context ``vacation'' $\rightarrow$ ``walk outside,'' people typically prefer \emph{``will''} for the second one. Similarly, any system which correctly answers this example for the right reason would need to incorporate TCS in its reasoning.

%\bzch{\sout{This example illustrates the importance of TCS in the reasoning process.}}

% There is a fair amount of reporting biases when it comes to commonsense understanding from text. Free texts are unlikely to report the most commonsense information, so it would be difficult for a model to learn them.

\begin{table}[h]
    \centering
\resizebox{1.0\hsize}{!}{
    \small 
	\begin{tabular}{|p{8cm}|}
		\hline
		\textbf{Example~\addexctr\label{ex:tcs}: choosing from {\em ``will'' or ``will not''}}\\
		Dr. Porter is now (\idxevent{\addeventCtr\label{ev:vacation}}{taking}{black}) a vacation and \underline{\qquad\qquad} be able to see you soon.\\
		Dr. Porter is now (\idxevent{\addeventCtr\label{ev:walk}}{taking}{black}) a walk outside and \underline{\qquad\qquad} be able to see you soon.\\
		\hline
	\end{tabular}
}
\end{table}

\begin{figure}[tbp!]
    \centering
    \resizebox{1.0\hsize}{!}{
        \includegraphics[scale=0.15,trim=1.6cm 0cm 0cm 1.5cm, clip=false]{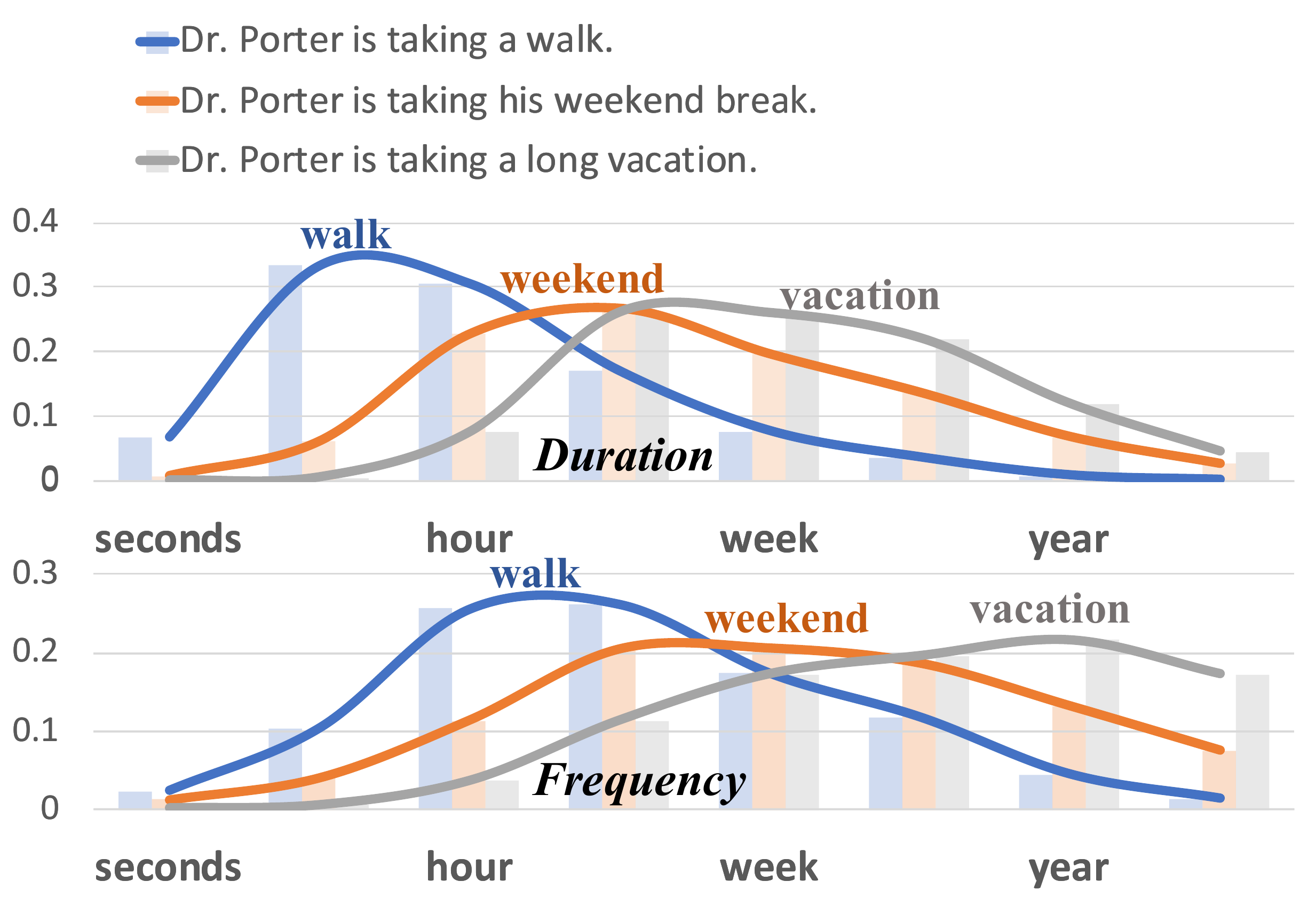}
    }
    \caption{
    Our model's predicted distributions of event \textbf{duration} and \textbf{frequency}. The model 
    is able to attend to contextual information and thus produce reasonable estimates. }
    % \qn{``fine-grained context'' is a bit handwavy to me} contexts and produce quality estimations.} 
    % \qn{it's not obvious why the predictions above are of ``quality.'' Like Daniel mentioned, the first impression people see this figure is that they are so smooth, which means they're too vague.}.
    % \danielk{ ... able to attend to contextual information and produce reasonable estimates.}
    %}
    \label{fig:predicted-dur-freq}
\end{figure}

\ignore{
\begin{figure}[tbp!]
    \centering
    \resizebox{1.0\hsize}{!}{
        \includegraphics[scale=0.23,trim=1.6cm 0cm 0cm 1.5cm, clip=false]{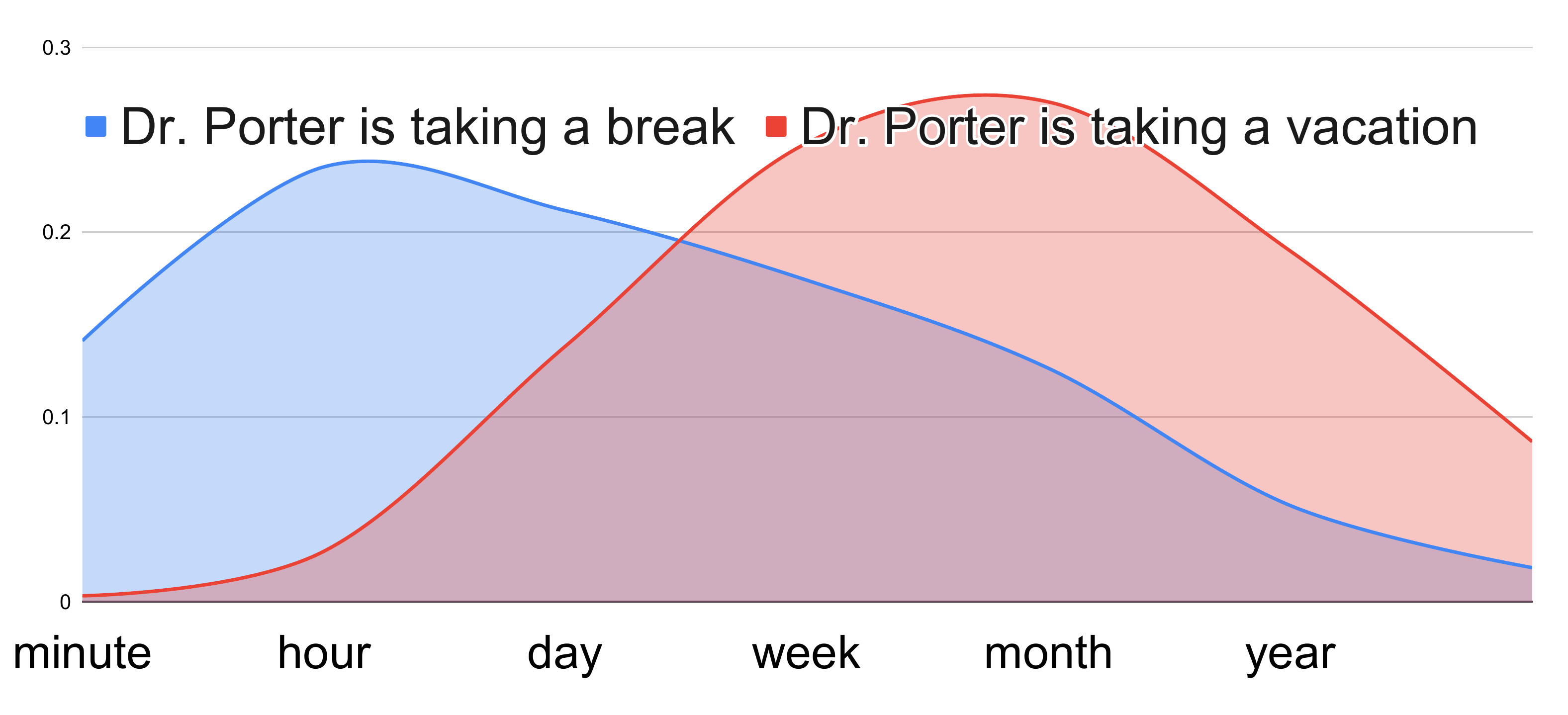}
    }
    \resizebox{1.0\hsize}{!}{
        \includegraphics[scale=0.23,trim=1.6cm 0cm 0cm 1.5cm, clip=false]{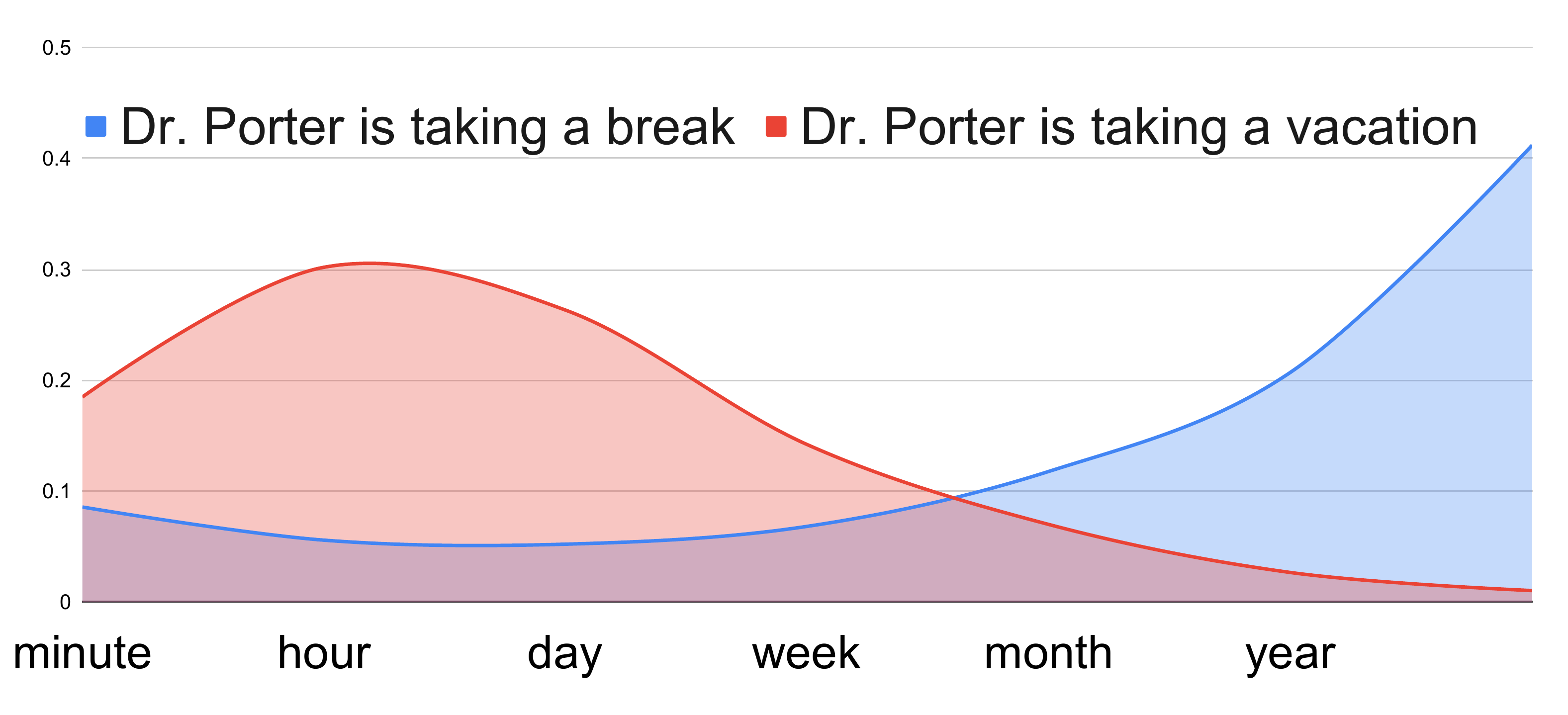}
    }
    
    \caption{The proposed model's prediction about the \textbf{duration} (top) and \textbf{frequency} (bottom) of two events.
    \danielk{fix the cropped figures.}
    }
    \label{fig:predicted-contextual-duration}
\end{figure}
}

% \danielk{
% I think somewhere we have to clarify why Fig1/Fig2 are smooth. 
% } \bz{I think they explain themselves now. Fig 2 has some style issue: it overlaps with text in the previous section.}

Acquiring the multiple dimensions of TCS (e.g., \textit{duration} and \textit{frequency})
%\bzch{in various dimensions such as \textit{duration} and \textit{frequency}} 
is challenging.
%due to two reporting biases.
%First, temporal information is often implicit. 
As shown in Example~\ref{ex:tcs}, the duration of ``taking a vacation'' and ``taking a walk'' are not expressed explicitly,
%in text,  
\bzch{so that systems are required} to read between the lines to support the inference.
%and rely on its common sense knowledge to infer that. 
A pre-trained language model may not handle this issue well, as it cannot identify the TCS dimensions in temporal mentions and effectively learn from them. As a result, it cannot generalize well to similar events without explicit temporal mentions.
To handle this problem, we design syntactic rules that can collect a vast amount of explicit mentions of TCS from unannotated corpus such as Gigaword~\cite{graff2003english} (\S\ref{subsec:cheap:supervision}). We use this data to pre-train our model so that it
%is able to 
distinguishes different dimensions.

%The second reporting bias exists in these explicit mentions.
A second challenge occurs when \bzch{the text is highlighting rare and special cases}.
\bzch{As a result, temporal mentions in natural text} %in this case %way
may follow a distorted distribution in which certain kinds of ``common'' events are under-represented.
For instance, we may rarely see mentions of 
% ``I opened the window in 3 seconds'' in text (the bias), but we see more often phrase like ``I opened the window five minutes ago'', indicating an \emph{upper-bound} of the event (i.e., ``opening the window'' doesn't take more than a minute).
``I opened the door in 3 seconds,'' but we may see ``it took me an hour to open this door" in text.  
To overcome this challenge, we exploit the joint relationship among 
%these 
temporal dimensions. Although we rarely observe the \emph{true} duration of ``opening the door'' in free-form text,
we may see %more often 
phrases like ``I opened my door during the fire alarm,'' providing an upper-bound to the duration of the event (i.e., ``opening the door'' does not take longer than the alarm.) 
We believe that we are the first to exploit such phenomena among %the
temporal dimensions. 
This paper studies several important dimensions of TCS inference: 
\emph{duration} (how long an event takes), \emph{frequency} (how often an event occurs)  
and
\emph{typical time} (when an event typically happens).\footnote{E.g., typical time in a day (the morning), typical day of a week (on Sunday), and typical time of a year (summer).}
% \bzch{\sout{and \emph{temporal ordering} (typical order of events).}}
% \danielk{the previous sentence is accurate? do we study 3 dimensions or 4? In S3 we say 3.} \bz{I changed to 3 dimensions.}
% Previous work has investigated some of these aspects, either explicitly or implicitly (e.g., duration \cite{divyekhilnani2011using,Williams12} and ordering \cite{ChklovskiPa04,NWPR18}), but none of them have defined or studied all aspects of temporal commonsense in a unified framework. 
As a highlight, Fig.~\ref{fig:predicted-dur-freq} shows the distributions (over time units) we predict for the {\em duration} and {\em frequency} 
%of our system for 
of three events.
% For instance, 
% Fig.~\ref{fig:predicted-dur-freq} shows the output of our system when predicting the \emph{duration} and \emph{frequency} of three related events in Example~\ref{ex:tcs}. 
We can see that
% The result is a distribution that indicates that
``taking a vacation'' lasts from days to months while ``taking a walk'' lasts from minutes to hours. 
% more details about the model: mined XX sentences, trained an LM, tested on X datasets
As shown, our model is able to produce different and sensible distributions for the ``take" event, depending on the context in which ``take" occurs.
% As compared to prior work \bz{Dan D pointed out that other works (say UDST) also distinguishes the context, so it might not be good if we claim 'unique'.}, it is a unique capability that our model produces reasonably different distributions for events in different contexts, although they are headed by the same verb ``take.''
% \dr{I think that it is enough to say: As shown, our model is able to produce different, and sensible, distributions for the ``take" event, depending on the context in which ``take" occurs.}
% in Fig.~\ref{fig:predicted-dur-freq}. 

\begin{figure}[htbp!]
    \centering
    \resizebox{1.0\hsize}{!}{
        \includegraphics[scale=0.14,trim=1.4cm 0.2cm 0cm 0.9cm, clip=false]{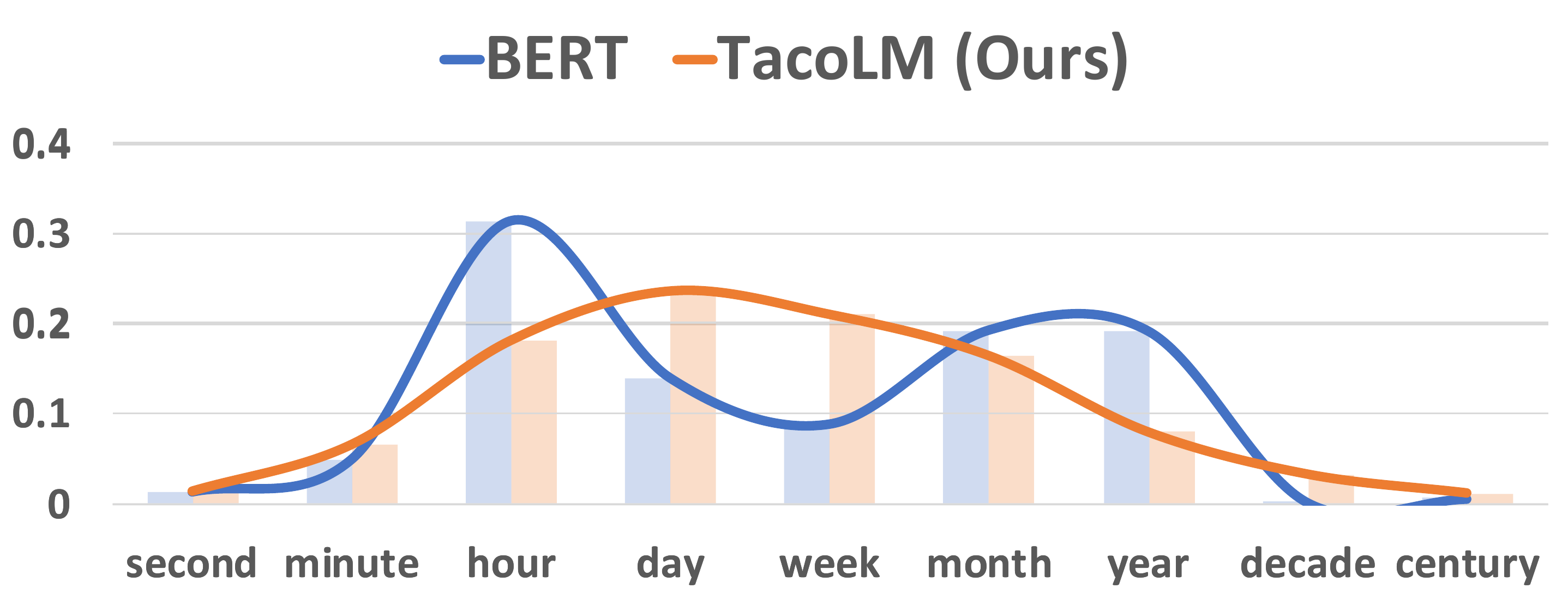}
    }
    \caption{
        The predictive distribution of two models (ours and vanilla BERT)
         for a set of events labeled with ``days" as their duration. Experiments show our model is about 40\% better on duration predictions. %indicated ``days'' as their duration 
        (RealNews corpus; details in \S\ref{subsec:intrinsic}).
        % Comparison of predicted label distributions on a set of events that has duration ``days'' 
    }
    \label{fig:predicted-distribution-comparison}
\end{figure}

% main point being: current models don’t utilize the [ordinal] relationship between temporal units. 
Our work builds upon pre-trained contextual language models~\cite{peters2018deep,devlin2018bert,liu2019roberta}.
However, a standard language modeling objective does not lead to a model that handles the two challenges %reporting biases
mentioned above; in addition, other systematic issues limit its ability to handle TCS.
% While such models are shown to be effective tools for a wide range of natural-language tasks, there are structural issues that limit their ability to handle temporal common sense. 
In particular, language models do not directly utilize the ordinal relationships among temporal units. For example, ``hours'' is longer than ``minutes,'' and ``minutes'' are longer than ``seconds.''\footnote{The relationship can be more complex. E.g., ``hours'' is closer to ``minutes'' than ``centuries'' is; days of a week forms a circle: ``Mon.'' is followed by ``Tue.'' and preceded by ``Sun.''} 
Fig.~\ref{fig:predicted-distribution-comparison} shows that BERT %produces a less meaningful 
does not produce a meaningful duration distribution for a set of events with a gold duration of ``day'' (extracted in \S\ref{subsec:cheap:supervision}). %On the contrary, 
Our proposed system, on the other hand, is able to utilize the ordinal relationships and produce uni-modal distributions around the correct labels in both Fig.~\ref{fig:predicted-dur-freq} and Fig.~\ref{fig:predicted-distribution-comparison} .
% \bz{The original version is in the "ignore" section below. I think it spends too much space on this issue, although it is not the highlight of this paper.}
\ignore{
Our proposed system directly encodes such external knowledge in the objective function of an augmented language model and is able to exploit these ordinal relations between temporal units during pre-training (\S\ref{subsec:soft_cross_entropy}).
% and also reflect it when making predictions. 
For instance, in Fig.~\ref{fig:predicted-dur-freq} our model generally produces uni-modal distribution around the correct labels. However, the \bzch{the predictive distributions \sout{output}} of BERT can often be multi-modal, indicating that it does not capture the ordinal relation between temporal units.
\bzch{\sout{Additionally,}} Fig.~\ref{fig:predicted-distribution-comparison} juxtaposes the predictive
% \danielk{
%    \sout{shows the aggregated prediction}
%    juxtaposes the predictive
% }
 distributions for our model and BERT~\cite{devlin2018bert} for 
 a set of events all with a gold duration of ``day'' \bzch{extracted in \S\ref{subsec:cheap:supervision}.}
Our model in aggregate produces uni-modal distribution around ``day,'' (which is where the gold labels are) while BERT model produces a less meaningful distribution.  
In \S\ref{subsec:intrinsic} we will see that our model significantly improves BERT in TCS understanding.  
% \danielk{talk about circularity (week days) too?}
% \danielk{emphasize that we are the first to learn three temporal dimensions \textbf{jointly}. }
}

% \vspace{-0.5em}

% Summary of shiny contributions: improved by X, some promise about releasing
\paragraph{Contributions.}
This work proposes an augmented pre-training for language models to improve their understanding of several important temporal phenomena. 
\bzch{We address two kinds of reporting biases by effectively acquiring weak supervision from free-form text and utilizing it to learn multiple temporal dimensions jointly.}
Our model %also 
incorporates
%keeps 
other desirable properties 
%about 
of
time  in its objective (ordinal relations between temporal phrases, the circularity of certain dimensions, etc.) to improve temporal modeling. 
Our experiments show 19\% relative improvement over BERT in intrinsic evaluations, and 5-10\% improvements in most extrinsic evaluations done on three time-related datasets.
Furthermore, the ablation study shows the value of each proposed component of our construction. Overall, this is the first work to incorporate a wide range of temporal phenomena within a contextual language model.
% \footnote{ Model to be released upon publication. } 

The rest of this paper is organized as follows. We distinguish our work with the prior work in \S\ref{sec:related_work}. The core of our construction, including extraction of cheap supervision from raw data and augmenting a language model objective function with temporal signals, is in \S\ref{sec:main_details} . We conclude by showing intrinsic and extrinsic experiments in \S\ref{sec:experiments}.

\section{Related Work}
\label{sec:related_work}
% \qn{common sense work: can borrow from relevant sections of McTACO paper}
% \qn{temporal work: can also borrow from relevant sections of McTACO paper}
% \qn{below is copied from mctaco paper. need to modify it.}
% \qn{McTACO, Yanai's work on numeracy, ``Do NLP models know numbers?''} --> cited it among other "physical" common sense works. s

% general commonsense  
Common sense has been a popular topic in recent years, and existing NLP works have mainly investigated the acquisition and evaluation of common sense reasoning in the physical world. These works include but are not limited to size, weight, and strength~\cite{bagherinezhad2016elephants,forbes2017verb,elazar2019large}, roundness and deliciousness~\cite{yang2018extracting}, and intensity~\cite{CWPAC18}. 
A handful of these works uses cheap supervision. For example, \citet{elazar2019large} recently proposed a general framework that discovers distributions of quantitative attributes (e.g., length, mass, speed, and duration) from explicit mentions (or co-occurrences) of these attributes in a large corpus. However, \citet{elazar2019large} restrict events to be verb tokens, while we handle verb phrases containing more detailed information (e.g., ``taking a vacation'' is very different from ``taking a break,'' although they share the same verb ``take''). Besides, there has been no report on the effectiveness of this method on temporal attributes.

On the other hand, {\em time} has long been an important research area in NLP. 
Prior works have focused on the extraction and normalization of temporal expressions~\cite{strotgen2010heideltime,angeli2012parsing,LADZ14,VashishthaVaWh19}, temporal relation extraction~\cite{NingFeRo17,NZFPR18,VashishthaVaWh19}, and timeline construction~\cite{LeeuwenbergMo18}. 
Recently, \mctaco{} \cite{ZKNR19} summarizes five types of TCS and the three temporal dimensions studied here are all in their proposal.\footnote{They additionally propose {\em typical order of events} and {\em stationarity} (whether a state holds for a very long time or indefinitely).}
% \emph{duration} (how long an event takes),  \emph{temporal ordering} (typical order of events), \emph{typical time} (e.g., people typically go to work in the morning), \emph{frequency} (how often an event occurs), and \emph{stationarity} (whether a state holds for a very long time or indefinitely).
\mctaco{} shows that modern NLU techniques are still a long way behind humans on TCS understanding, suggesting that further research on this topic is needed. 

There have been works on temporal common sense, such as event duration \cite{PanMuHo06,GCKKBJ11,williams2012extracting,VempalaBlPa18,VashishthaVaWh19}, typical temporal ordering~\cite{ChklovskiPa04,NFWR18,NWPR18}, and script learning (i.e., what happens next after certain events)~\cite{GranrothCl16,LiDiLi18,PengNiRo19}. 
Those on duration are highly relevant to this work.
\cite{PanMuHo06} annotates a subset of documents from TimeBank~\cite{PHSSGSRSDFo03} with ``less-than-one-day'' and ``more-than-one-day'' annotations and provides the first baseline system for this dataset.
\citet{VempalaBlPa18} significantly improve earlier work by using additional aspectual features for this task.
\citet{VashishthaVaWh19} annotate the UDS-T dataset with event duration annotations and propose a joint method that extracts both temporal relations and event durations. 
Our approach has two notable differences from this line of work. First, we work on duration, frequency, and typical time---jointly on three dimensions of TCS, while the works above only focused on duration. Second, we focus more on obtaining cheap supervision signals from unlabeled data, while these other works all have access to human annotations.
With respect to harnessing cheap supervision, \citet{williams2012extracting,GCKKBJ11} propose to mine web data using a collection of hand-designed query patterns. In contrast to our approach, they are based on counting instead of machine learning and cannot handle the contextualization of events.

\section{Temporally Focused Joint Learning with Minimal Supervision}
\label{sec:main_details}

In this section, we elaborate our approach to designing and pre-training \modelname, a time-aware language model.
% \qn{can you double check if this is the first time we mention the acronyms of \modelname{} and LM? If yes, I suggest ``...we elaborate our approach to further pre-train an existing language model (LM) such as BERT with an additional focus on time.''}
% \danielk{
%     \sout{our method that improves over previous language models pre-training for temporal common sense representation.}
%     % the discussion of "improves over previous language models" does not belong here; it belongs to the next section 
%     our approach to designing and pre-training a temporal-aware language model. 
% }

\subsection{Scope}

In this work, we focus on three major temporal dimensions of events, namely \textit{Duration}, \textit{Frequency} and \textit{Typical Time}. Here, \textit{Typical Time} means the typical occurring time of events during a day, day of a week, and month or season of a year.
We follow the same definition to each of the dimensions (also called properties) in~\citet{ZKNR19}.

\subsection{Joint Learning and Auxiliary Dimensions}
\label{sec:joint_learning_auxilary_dimensions}
As mentioned earlier, commonsense information extraction comes with the challenge of reporting biases. For example, people may not report the duration of ``opening the door,'' or the frequency of ``going to work.'' However, it is often possible to get supportive signals from other dimensions, as people mention ``going to work'' associated mostly with ``a day'' in a week, hence we may know the frequency of such an event. 

We argue that many temporal dimensions are interrelated and a joint learning scheme would suit this task. Beyond duration, frequency and typical time, we also introduce 
auxiliary dimensions that are not meant to be used by themselves
    % \sout{additional dimensions that we call auxiliary dimensions, which may not be useful by themselves,}
but will help the prediction of other dimensions. The auxiliary dimensions we define here are event \textit{Duration Upper-bound} and event \textit{Relative Hierarchy}. The former represents values that are upper-bounds to an event's duration but not necessarily the exact duration. The latter consists of two sub-relations, namely \emph{temporal ordering} and \emph{duration inclusion} of event-pairs.

\subsection{Cheap Supervision from Patterns}
\label{subsec:cheap:supervision}

\begin{figure}
    \centering
    \includegraphics[scale=0.28,trim=1.65cm 2.0cm 1.1cm 0.9cm, clip=false]{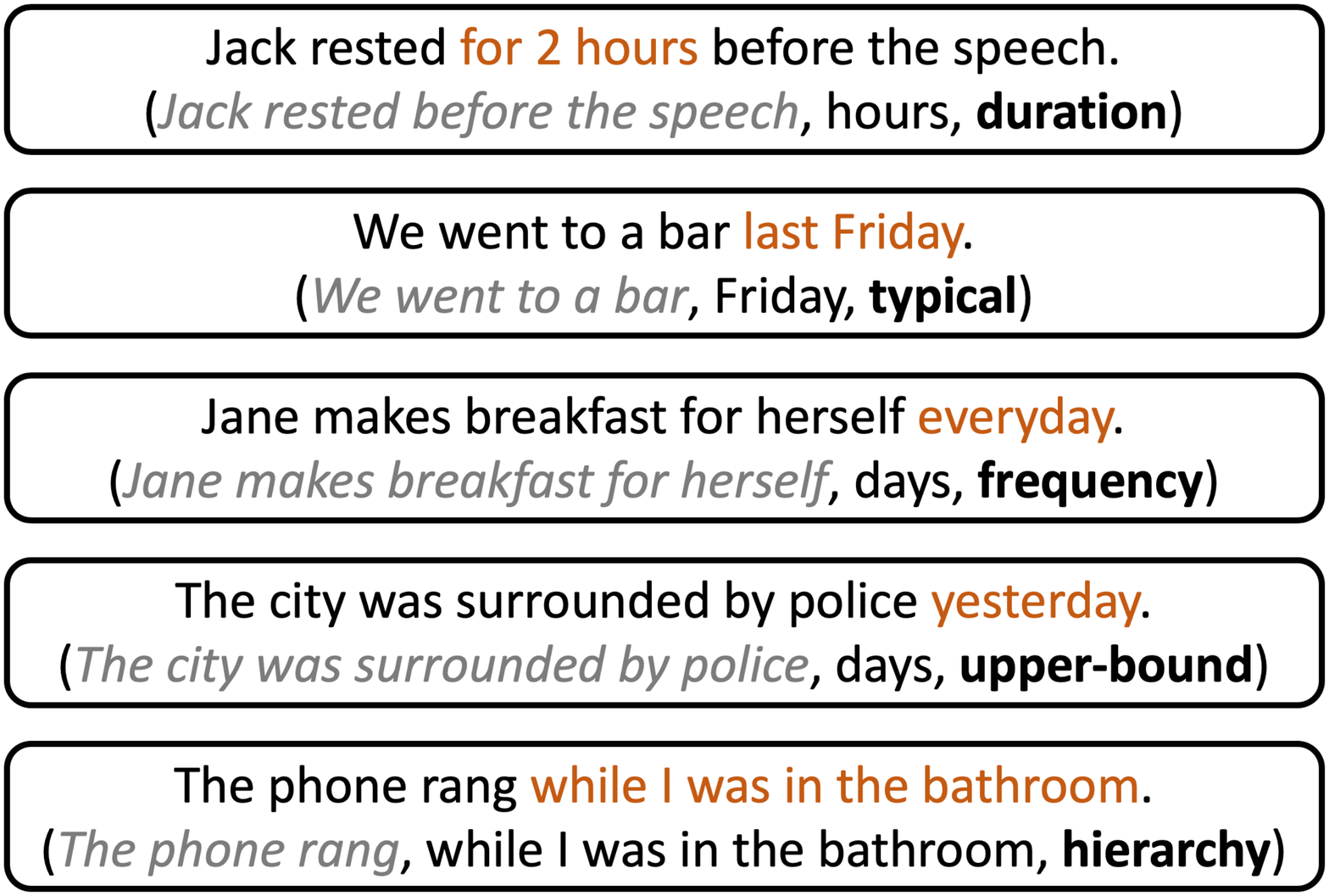}
    \caption{
        Examples of the extraction process for each temporal dimensions. 
        The temporal arguments are marked \orangetext{orange} and the result of the extraction are tuples of the form (\keywordCode{event,value,dimension}).
    }
    \label{fig:pattern-example}
\end{figure}

We collect a few pattern-based extraction rules based on SRL parses for each temporal dimension (including the auxiliary dimensions). We design the rules to have high precision, while not compromising too much on recall. 
We overcome the potential sparsity issue (and the resulting low recall problem) 
 by extracting from a massive amount of data.
%We do not measure the recall during the design as it was secondary for us. \sihao{why? Maybe add some explanations here?}
Fig.~\ref{fig:pattern-example} provides some examples of the input/output for each dimension, as we describe the specific extraction process below.

We first process the entire Gigaword~\cite{graff2003english} corpus and use AllenNLP's SRL model~\cite{gardner2018allennlp,shi2019simple} to annotate all sentences. We extract the ones that contain at least one temporal argument (i.e., the \emph{arg-tmp} constituent of SRL annotations) and use textual patterns to categorize each sentence
% \danielk{\sout{classify} categorize the sentence} each of these arguments 
into a corresponding dimension with respect to an associated verb.
These patterns are inspired by earlier works and are extensively improved  with iterative manual error analysis. 
The rest of this section is devoted to explaining the key design ideas used for these patterns. 

\noindent
\textbf{Duration.} We check if the temporal argument starts with ``for,'' extract the numerical value and the temporal unit word, and normalize them into the nearest unit among the nine units in our scope: (``second,'' ``minute,'' ``hour,'' ``day,'' ``week,'' ``month,'' ``year,'' ``decade,'' ``century.'')  
We ignore particular phrases such as ``for a second chance'' where the semantic of ``second'' is not temporal related.
% \danielk{clarify this; potentially mention an example where ``for a second'' does not convey a proper duration. Also, replace ``irrelevant'' with a better phrasing.} \bz{better?}
We found that ``for'' is the only high-precision preposition that indicates exact values of duration. 

\noindent
\textbf{Frequency.} Such temporal arguments are usually composed of a duration phrase and a numerical head (e.g., ``four times per'') indicating the frequency within the duration (e.g., ``week''). 
% \danielk{include an example here; and indicate the frequency and duration.} \bz{better?}
Thus, we check for multiple keywords that indicate the start of a frequency expression, including ``every,'' ``per,'' ``once,'' \dots ``times.'' If so, we extract the duration value as well as the numerical head's value. 
% \danielk{``the number of times value'' is not clear.}
We ignore any temporal phrases that contain ``when'' since they often convey semantics that does not fit any of our temporal categories; e.g., ``when everyday life...'' is not describing the frequency of the corresponding verb.
We represent the frequency with duration $\mathbf{d}$, with a definition of
occurring \emph{once} every $\mathbf{d}$ elapses. For example, the frequency of ``four times per week'' is represented as ``1.75 days.'' Similarly, we normalize them into the nearest unit among the nine duration units described above, and ``1.75 days'' is extracted as ``days.''
% \danielk{
% Rephrasing of the previous two sentences, if I understand it: 
% We normalize the extracted frequency values by representing  them based on a neighboring (and smaller) unit of duration. For example, the frequency of ``four times per week'' is represented as with the unit of duration of ``days''.
% } \bz{I think your version misses the point on representing frequency with duration. I rephrased my version, lmk how it looks.}

\noindent
\textbf{Typical Time.} We pre-define a list of typical time keywords, including the \emph{time of day} (e.g., ``morning'' etc.), \emph{time of week} (e.g., ``Monday'' etc.), \emph{month} (e.g., ``January'' etc.) and \emph{season} (e.g., ``winter'' etc.) We check if any of the typical time keywords appear in the temporal argument and verify if the temporal argument is, in fact, describing the time of occurrence. This is done by filtering out the temporal arguments that contain a set of invalid prepositions, including ``until,'' ``since,'' ``following,'' since such keywords often do not indicate the actual time of occurrence. 

\noindent
\textbf{Duration Upper-bound.} 
Many temporal arguments describe the duration \emph{upper-bound} instead of the exact duration value. For example, as described in \cite{GCKKBJ11}, ``did \keywordCode{[activity]} yesterday'' indicates something that happened within a day. We extend the set of patterns to include ``in \keywordCode{[temporal expression]}'' or keywords such as ``next'' (e.g., ``the next day''), ``last'' (e.g., ``last week''), ``previous'' (e.g., ``previous month''), or ``recent'' (e.g., ``recent years''). 
%  \danielk{provide an example to clarify how these keywords express a duration upper-bound?} \bz{does this work?}
 We normalize the values into the same label set of the nine unit words as the duration dimension.
% \danielk{not clear where is ``above''. Also, this is the first time we mention ``9''.} \bz{we mentioned 9 in the Duration section too.}

\noindent
\textbf{Event Relative Hierarchy.} 
\bzch{A system can learn about an event with comparisons to other ones, as we show in \S\ref{sec:intro}. To acquire hierarchical relationships between events, we} check whether the SRL temporal argument starts with a keyword that indicates a relation
between the main event and another event phrase. We consider five such keywords, namely ``before,'' ``after,'' ``during,'' ``while'' and ``when.'' 
We use these keywords to label the relative relationship between the two events. 
Here, we assume that ``during'' and ``while'' are the same, which indicates that the main event is not longer than the one in the argument. 
% We also note that ``while'' has a different sense (similar to ``whereas'') and we rely on the SRL parser to correctly identify them as non-temporal arguments.
Note that certain keywords might have meanings that do not suggest temporal relationships (e.g., ``while'' has a different sense similar to ``whereas.'')
We rely on SRL annotations to identify the appropriate sense of the keywords.
We use the temporal keyword as labels, but keep the entire event phrase in the SRL temporal argument for later use in \S\ref{subsec:language_model}.
% \danielk{
% My rephrasing of the last sentence: 
% Note that certain keywords might have meanings do not indicate temporal relationships between events (e.g, ``while'' has a different sense similar to ``whereas.'')
% We rely on SRL annotations to identify the appropriate sense of the z5 % % % keywords.
% }
%\sihao{Note that the conjunction ``while'' has a different word sense closer to ``whereas''. In these cases, we assume that the SRL parser correctly identifies....}

\ignore{\danielk{
Do you want to reference ``Allen's interval agebra'' here? 
\url{https://en.wikipedia.org/wiki/Allen\%27s_interval_algebra}
\cite{allen1983maintaining}
}}

\begin{figure}[h]
    \centering
    \includegraphics[scale=0.23,trim=0.4cm 0.7cm 0cm 0cm]{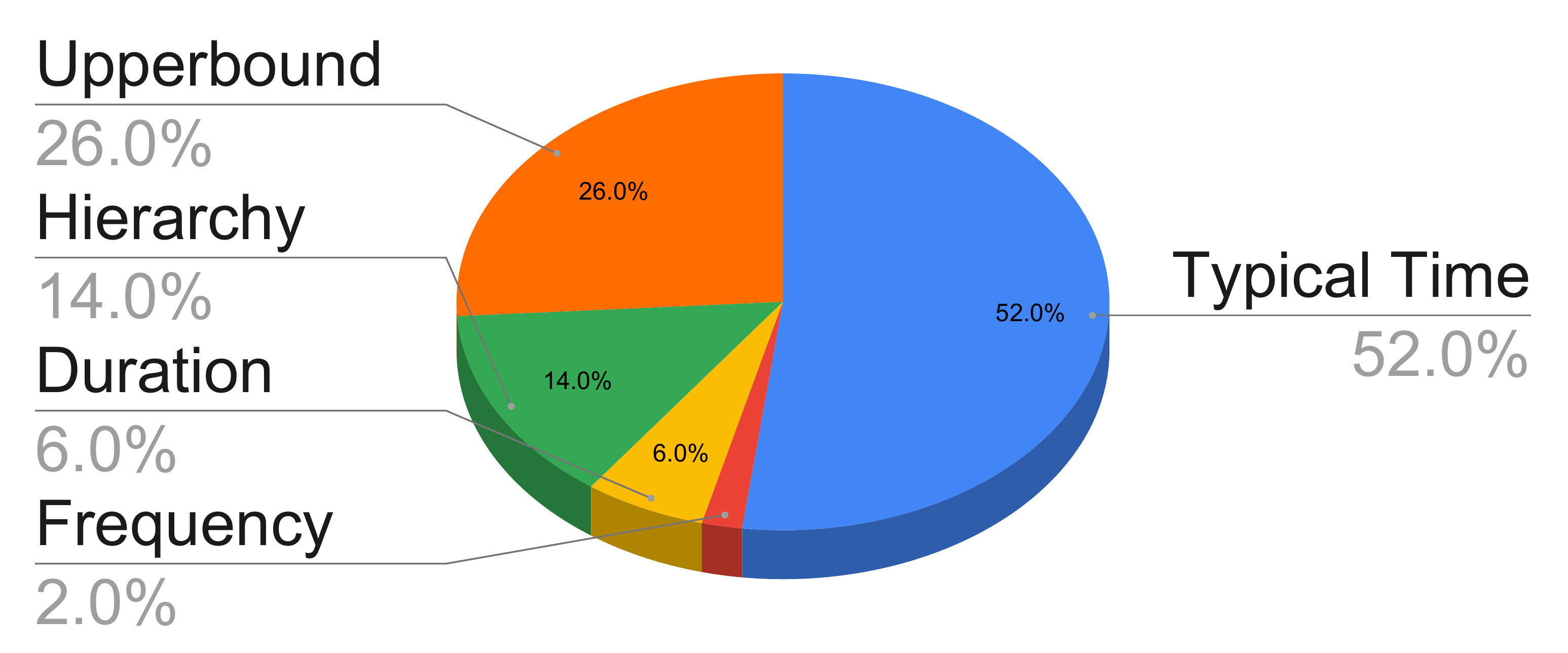}
    \caption{The distributions of different temporal dimensions in the collected data.}
    \label{tab:distribution-pattern}
\end{figure}

\noindent
\textbf{Resulting data.}
We collect 25 million instances that are successfully parsed into one of our temporal dimensions
% \danielk{with at least one temporal argument,}  
% out of 76 million \bzch{\sout{raw sentences} sentences with at least one temporal keywords}
% \danielk{\sout{with at least one temporal argument}} 
from the entire Gigaword corpus~\cite{graff2003english}. Each instance is in the form of (\keywordCode{event,value,dimension}) tuples (Fig.~\ref{fig:pattern-example}), with a dimension distribution shown in Fig.~\ref{tab:distribution-pattern}. 
For all events, we remove the related temporal argument so that it does not contain direct information about the dimension or the value. For example, as shown in Fig.~\ref{fig:pattern-example}, ``for 2 hours'' is removed, and only ``Jack rested before the speech'' is kept so that the target duration does not present in the event.
Note that \keywordCode{value} is also called and used as ``label'' in later contexts related to classification tasks.

% \begin{table}[h]
% \small
% \centering
% \begin{adjustbox}{width=0.48\textwidth}
%     \begin{tabular}{l|c|c|c|c}
%         \toprule 
%         Duration & Frequency & Typical Time & Upperbound & Relationship \\
%         \cmidrule(lr){1-1}                              \cmidrule(lr){2-2}  \cmidrule(lr){3-3} \cmidrule(lr){4-4} \cmidrule(lr){5-5}
%         6\% & 2\% & 52\% & 26\% & 14\% \\
%         \bottomrule 
%     \end{tabular}
% \end{adjustbox}
% \caption{\small Distributions of pattern collected data}
% \label{tab:distribution-pattern}
% \end{table}

%\danielk{general comment: I think you use `label' and `value' interchangeably. }

\subsection{Soft Cross-Entropy Objective for Ordinal Classification}
\label{subsec:soft_cross_entropy}
\ignore{
Many of the temporal keywords exist ordinal relationships. 
\danielk{not clear; how about something like this? 
The temporal value of an event in one dimension is often tied/related/connected to the the temporal value of the same event in a other dimensions.
}}
The temporal values in one dimension are naturally related to each other via a certain ordering and appropriate distance measures. To account for and utilize this external knowledge, we use a soft cross-entropy to encourage predictions that are aligned with the external knowledge. 

% \danielk{DOES THIS ADD ANY VALUE?} \bz{removed.}
% We create a soft target vector $\mathbf{y}$ to represent target labels' desired probabilities that align with the external knowledge. Thus, $\mathbf{y}$ has non-zero values and sum to 1.0. Consider $\mathbf{x}$ as a system's output logits across labels, we express our soft objective as the following: 
% $$\hat{\mathbf{x}} = \log(\text{softmax}(\mathbf{x}))$$
% $$\text{loss} = -\hat{\mathbf{x}}^\top  \mathbf{y}$$

Consider $\mathbf{x}$ as a system's output logits across labels, and we express our soft loss function as follows: 
\begin{equation}
    \ell = - \sum_{i \in D}  \mathbf{y}_i^\top \log(\text{softmax}(\mathbf{x}_i)),
\end{equation}
where $D$ is the instances in the training data and  $\mathbf{y}$ represent the degree to which the target labels align with the external knowledge. 
Thus, $\mathbf{y}$ is a probability vector, i.e., has non-zero values and sum to 1.0.

Now we describe how we construct $\mathbf{y}$ to apply the aforementioned external knowledge. \emph{Duration}, \emph{Frequency}, and \emph{Upper-bound} take the same set of labels of duration units. 
We first define a function $\text{logsec(.)}$ which takes a unit and normalizes it to its logarithmic value in ``seconds'' (e.g., ``minute'' $\rightarrow$ 60 $\rightarrow$ 4.1).
For each instance in these dimensions, with an observed gold label $g$, we assume a normal distribution with a mean value of $\mu= \text{logsec}(g)$ and a fixed standard deviation of $\sigma=4$. 
Then, we construct $\mathbf{y}$ so that, 
\begin{equation}
\mathbf{y}[i] = \frac{1}{{\sigma \sqrt {2\pi } }}e^{{{ - \left( {\text{logsec}(l) - \mu } \right)^2 } \mathord{\left/ {\vphantom {{ - \left( {\text{logsec}(l) - \mu } \right)^2 } {2\sigma ^2 }}} \right. \kern-\nulldelimiterspace} {2\sigma ^2 }}}
\end{equation}
where $l$ is the $i^{th}$ label. We apply softmax at the end to ensure $\mathbf{y}$ sums to 1.

% \danielk{
% The stylistic formatting of the dimension names are not consistent. For example, in the previous paragraph we start the sames with capital letters and make them italicized (e.g., \emph{Frequency}). But in the next paragraph it's written in a plain style (``typical time''.)
% } \bz{changed to textit below. I think it's different when we are referring to a name of a dimension, or the word's actual meaning.}

For \textit{typical time}, the labels are placed with approximately equal distances in a circular fashion. 
For example, ``Monday'' is before ``Tuesday'' and after ``Sunday.'' 
We assume adjacent units have a distance of 1, and we generate $\mathbf{y}$ based on a Gaussian distribution with a standard deviation of $0.5$. 
In other words, we assume the two immediate neighbors of a gold label are reasonably possible.

For \textit{hierarchy}, we construct $\mathbf{y}$ as a one-hot vector where only the gold label has a value of 1, and the rest are zeroes.

\subsection{Sequential Language Modeling}
\label{subsec:language_model}
Our goal is to build a model that is able to predict temporal labels (values) given events and dimensions. Instead of building a classification layer on top of a pre-trained model, we follow previous work~\cite{Huang2019GlossBERTBF} and place the label into the input sequence. We mask the label in the sequence and use the masked token prediction objective as the classification objective. To produce more general representations, we also keep the temporal label and mask the event tokens instead at a certain probability, so that we are able to maximize both \prob{\keywordCode{Tmp-Label} | \keywordCode{Event}} and \prob{\keywordCode{Event} | \keywordCode{Tmp-Label}} in the same learning process, where \keywordCode{Tmp-Label} refers to the temporal label associated with the event.
% \danielk{How about ``temporal label'' instead of ``knowledge''? If so, change Arg-Tmp to Tmp-Label, too.}

\ignore{\danielk{why not maximizing both at the same time?} \bz{You mean masking events and temporal values at the same time? I think because I wanted to maximize the capabilities of predicting temporal values given events.} \danielk{I see. Maybe if you write down the objective more explicitly it could help it clarify how each of these functions help accomplish the task.}}

%\danielk{in the following paragraph, in a couple of places you use ``vocabulary'' to mean ``lexicon entry''. Better to clarify so that there is no confusion. }

Specifically, we use the reserved ``unused'' tokens in BERT-base model lexicon to construct a 1-to-1 mapping from every value in every dimension to the new vocabulary. We choose not to use the existing representations for temporal terms that are already included in BERT's ``in-use'' lexicon, such as ``minutes'' or ``weeks,'' because these keywords have different temporal semantics in different dimensions. Instead, we assign unique and separate lexicon entries to different values in different dimensions, even though the values may share the same surfaces.
\ignore{
vocabulary that represents different \emph{dimensions} and \emph{values} of temporal properties discussed earlier. 
\danielk{
    explain why you need to augment the lexicon: doesn't BERT already contain entries like 'Sunday', 'morning', etc? 
}\bz{I had the next sentence for that purpose. Is it confusing?}
Note here that although some values have the same surface, such as ``minutes'' in duration and frequency, they are assigned different token representations, as they have different semantics by describing different temporal dimensions. 
\danielk{
    are you saying that you have multiple ``minute''s? (like one for duration and one for frequency? 
} \bz{Yes}
\danielk{
Maybe try to say it directly?: 
The temporal terms that could describe multiple dimensions, have multiple representatives in the lexicon, each corresponding to a distinct temporal dimension. For instance ``minutes'' could refer to either duration and frequency. 
}
}
%We use special markers to separate the temporal value with the contexts, as well as a marker that is placed on the left of the main event's verb (extracted by SRL). 
Consider each $($\keywordCode{event,value,dimension}$)$ tuple, we map \keywordCode{value} and \keywordCode{dimension} to their new vocabularies \keywordCode{[Val]} and \keywordCode{[Dim]}, and we use [\keywordCode{W}$_1$, \keywordCode{W}$_2$, $\hdots$, \keywordCode{W}$_n$] to represent the tokens in the sentence, and \keywordCode{W}$_{verb}$ the event verb anchor from SRL. 
We now form a sequence [\keywordCode{W}$_1$, \keywordCode{W}$_2$,$\hdots$\keywordCode{[Vrb]}, \keywordCode{W}$_{verb}$, $\hdots$\keywordCode{W}$_n$, \keywordCode{[SEP]}, \keywordCode{[Vrb]}, \keywordCode{[Dim]}, \keywordCode{[Val]}, \keywordCode{[Arg-Tmp-Event]}], where \keywordCode{[Vrb]} is a marker token that is the same across all instances. \keywordCode{[Arg-Tmp-Event]} is the event phrase in the SRL temporal argument, as described in \textit{hierarchy}. \keywordCode{[Arg-Tmp-Event]} is empty for all dimensions other than \textit{hierarchy}.
% \danielk{I can't remember why we didn't use \keywordCode{w} instead of \keywordCode{Event}, given that these are words.} \bz{good point, changed}

We mask \keywordCode{[Val]} with probability $p_{mask}$ and \keywordCode{[Dim]} with probability $p_{dim}$. We individually mask each event tokens with probability $p_{event}$ when we do not mask \keywordCode{[Val]} nor \keywordCode{[Dim]}.
% \danielk{
% Odd wording: the first part of the sentence active ("We mask") and the latter part is passive ("is not masked"). 
% }
Soft cross-entropy is used when predicting \keywordCode{[Val]}, and a regular Cross-entropy is used for other tokens. \bzch{We use the pre-trained token-recovery layer, and follow BERT's setting to randomly keep a token's surface or change it to noise during recovery.}

In the experiments, we explore a set of configurations of the system. 
We explore the effect of having only one sentence or the two additional neighboring sentences as input contexts.
We also experiment with all-event-masking, where we mask tokens in the event with a much higher probability. The goal of this masking scheme is to reduce the predictability of event tokens based on other event tokens to alleviate prior biases and focus more on the temporal argument. For example, BERT predicts ``coffee'' for the \keywordCode{[MASK]} in ``I had a cup of \keywordCode{[MASK]} this evening'' because of the strong prior of ``cup of.'' By masking more tokens in the event, the remaining ones will be more conditioned to the temporal cue.
\ignore{\qn{did you mention ``joint'' in this section? also, does ``joint'' deserve a dedicated section?}}

\subsection{Label Weight Adjustment}
\label{subsec:label_weight}
The label imbalance in the training data largely hinders our goal, as we should not assume a prior distribution as expressed in natural language.
% \danielk{The label impalance in the training data ... }
% \danielk{the sentence is under-specified; rework, cite an ``imbalance'' paper as well}
For example, ``seconds'' appears around ten times less than ``years'' in the data we collected for \textit{duration}, leading to a biased model. We use weight adjustment to fix this. Specifically, we apply weight adjustment to the total loss with a weight factor calculated as the observed label's count relative to the number of all instances.
% observed label's occurrence relative to the average count of its dimension.

%Different units within the same dimension are not uniformly distributed in free text, yet human makes no such prior assumption when asked to predict values in a given dimension. For example, ``seconds'' appear around 10 times less than ``years'' for duration. To fix this, we proposed a weight adjustment. Specifically, we sample the distributions in the training data, and apply a weight adjustment to the loss (both when predicting the temporal value and predicting the event when the value is not masked.) The weight factor is calculated as the number of a label's occurrence relative to the average number in the same dimension. 

\section{Experiments}
\label{sec:experiments}

\subsection{Variations and Settings}
We experiment with several variants of the proposed system to study the effect of each change.

\noindent
\textbf{Input Size.} A model with three input sentences (including the event sentence's left/right neighbors) are labeled with \keywordText{MS}. Non \keywordText{MS} models use only one sentence in which the event occurs.
% a model that uses single sentences is labeled with \keywordText{SS}. 

\noindent
\textbf{All Event Masking.} A model with $p_{event}=0.6$ is labeled as \keywordText{AM}, and  $p_{event}=0.15$ otherwise.

\noindent
\textbf{Final Model.} Our final model includes all auxiliary dimensions (\keywordText{AUX}) (mentioned in \S\ref{sec:joint_learning_auxilary_dimensions}), uses soft cross-entropy loss (\keywordText{SL}) and applies weight adjustment (\keywordText{ADJ}) (mentioned in \S\ref{subsec:label_weight}). We study each changes' effect by ablating them individually. 

To deal with the skew present in the training data (\S\ref{sec:main_details}), we down-sample to ensure roughly the same occurrences of each dimension (except for frequency because of its low quantity). As a result, 4.3 million sentences were used in pre-training (down-sampled from 25 million mined sentences). We employ a learning rate of 2e-5 with 3 epochs and set $p_{mask}=0.6$ and $p_{dim}=0.1$. Other parameters are the same as those of the BERT base model. We use epoch 2's model for extrinsic evaluations to favor generalization, and epoch 3's model for intrinsic evaluations as it achieves the best performance across tasks.

\subsection{Intrinsic Evaluation}
\label{subsec:intrinsic}
We evaluate our method on the temporal value recovery task,
where the inputs are a sentence representing the event, an index to the event's verb, and a target dimension. The goal is to recover the temporal value of the given event in the given dimension. 

\noindent
\textbf{Datasets.} To ensure a fair comparison, we sample instances from a new corpus RealNews~\cite{zellers2019neuralfakenews} that have no document overlap with our pre-training data and, at the same time making the data not strictly in-domain. We apply the same pattern extraction process mentioned in \S\ref{subsec:cheap:supervision} on the new data and collect instances that are uniformly distributed across dimensions and values. In addition, we ask annotators on Mechanical Turk to filter out the events that cannot be recovered by common sense. For example, ``I brush my teeth [Month]'' will be discarded because all candidate answers are approximately uniformly distributed so that one cannot identify a subgroup of labels to be more likely. 

Specifically, we ask one annotator to select from 4 choices regarding each (event, temporal value) tuple. The choices are 1) the event is unclear and abstract; 2) the event has a uniform distribution across all labels within the dimension; 3) the given label is one of the top 25\% choices among all other labels within the dimension and 4) the given label is not very likely. We keep the instances for which the annotator selects option 3), verifying that the label is a very likely choice for the given dimension. For the RealNews corpus, we annotate 1,774 events that are roughly uniformly distributed across dimensions and labels, among which 300 events are preserved.

We also apply the same process to UDST dataset. We find the majority of the original annotation to be unsuitable, 
as there are many annotations to events that are seemingly undecidable by common sense. We first apply an initial filtering by using only events of which the anchor word is a verb and require all existing annotations from \cite{VashishthaVaWh19} of the same instance to have an average distance less than two units. We then use our method to annotate 1,047 events, and eventually, 142 instances are left.

\noindent
\textbf{Systems.} In both datasets, we compare our proposed system with BERT. To use BERT's predictions on temporal values without supervision, we artificially add prepositions querying the target dimension as well as a masked token right after the verb. For example, ``I ran to the campus'' will be transformed as ``I ran for 1 \keywordCode{[MASK]} to the campus''. The specific prepositions added are ``for 1'' (duration), ``every'' (frequency), ``in the'' (time of the day), ``on'' (week), ``in'' (month), and ``in'' (season). We then rank the temporal keywords (singular) in the given dimension according to the masked token's predictions.
% \begin{itemize}
%     \item duration: for
%     \item frequency: every
%     \item time of the day: in the
%     \item week: on
%     \item month: in
%     \item season: in
% \end{itemize}
For our model, we follow the sequence formulation described above, recover and rank the masked \keywordCode{[Val]} token.

% \danielk{Suggestion: rename ``TmpBERT'' to ``BERT++''} \bz{done}

In addition, we also compare with a baseline system called \emph{BERT + naive finetune}, which is BERT fine-tuned on the same pre-training data we used for our proposed models, with a higher probability of masking a temporally related keyword (i.e., all values we used in all dimensions). Unlike our model, we only use soft cross-entropy loss and do not distinguish the dimensions each keyword is expressing. 

\noindent
\textbf{Metrics.} Following \citet{VashishthaVaWh19}, we employ a metric ``distance'' that measures the rank difference between a system's top prediction and the gold label with respect to an ordered label set. For duration and frequency where values are in a one-directional order, we use the absolute difference of the label ranks. For other dimensions where the labels are in circular relationships, we use the minimal distance between two labels in both directions, so that ``January'' will have a distance 1 with ``December.'' This is similar to an \emph{MAE} metric, and we report the averaged number across instances.

\begin{figure}[h]
\centering
% \danielk{commented out for speed}
\includegraphics[scale=0.33]{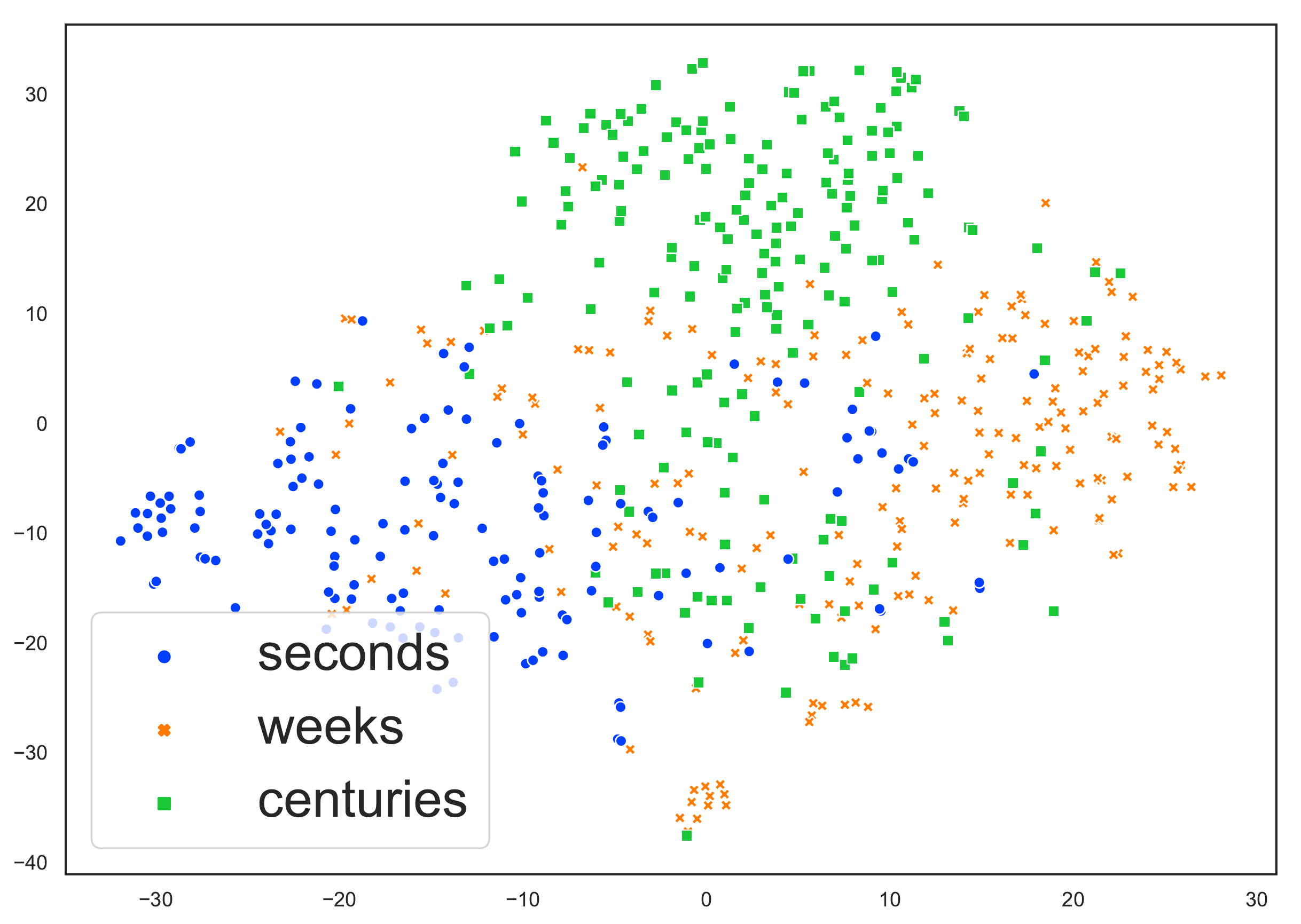}
\vspace{-1.5em}
\caption{
Representations of events (whose durations were labeled as seconds, weeks, or centuries) obtained from the original BERT base model.
% \qnch{Representations of events (whose durations were labeled as seconds, weeks, or centuries) obtained from the original BERT (base/large?). We used t-SNE visualization.}
% \qn{can we make the legends larger in Fig.5\&6? also, can we make the background white?} \bz{I think saying we used t-SNE is repetitive?}
}
\label{fig:bert-embedding-visualization}
% \danielk{commented out for speed}
\vspace{1em}
\includegraphics[scale=0.33]{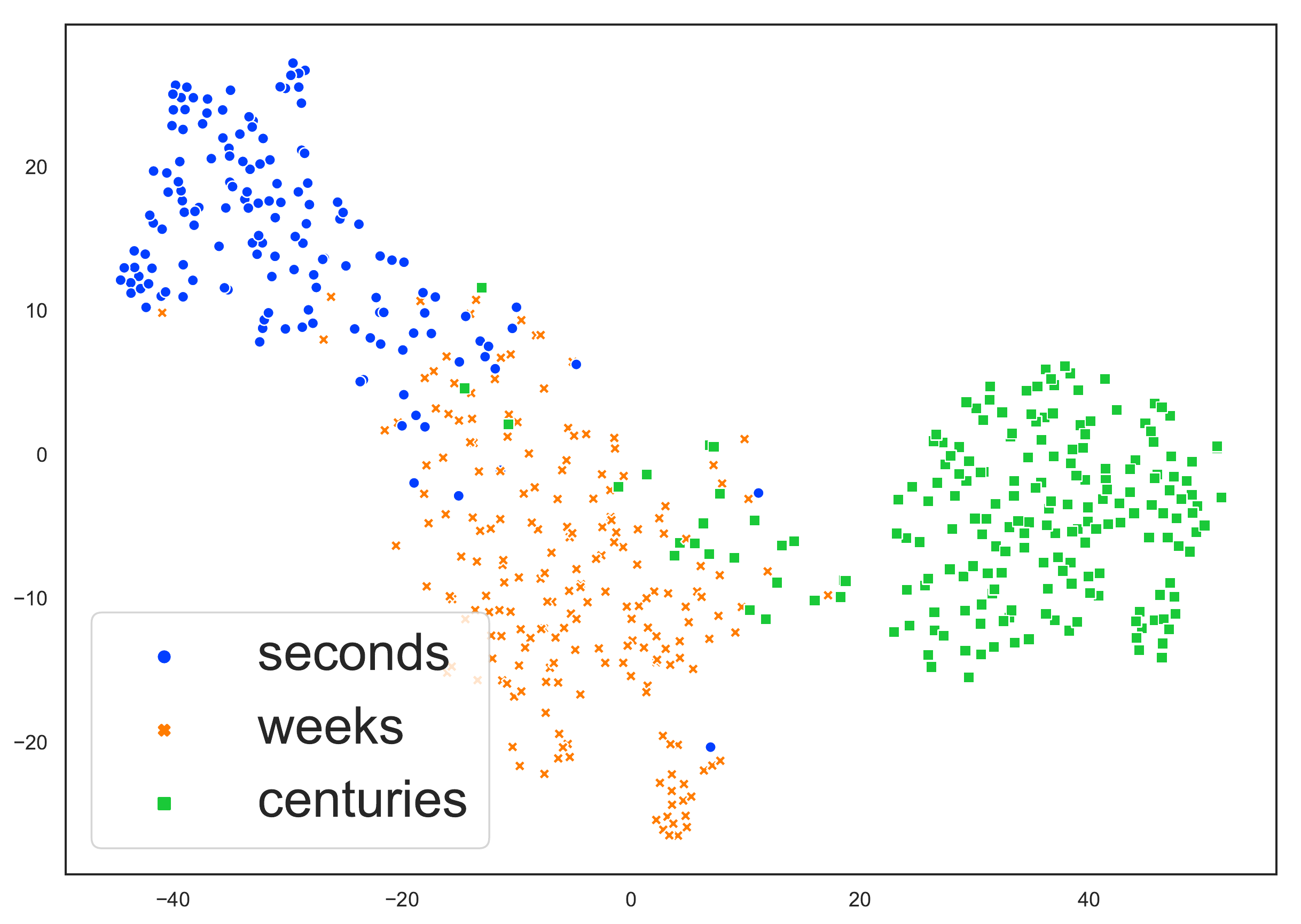}
\vspace{-1.5em}
\caption{
% \bzch{Our proposed embedding visualization of events with labeled duration of ``seconds'' ``weeks'' or ``centuries''}
Representations of the same set of events as in Fig.~\ref{fig:bert-embedding-visualization} obtained from the proposed method.
}
\label{fig:ours-embedding-visualization}
\end{figure}

\begin{table*}[]
\centering
\begin{adjustbox}{width=0.73\textwidth}
\begin{tabular}{l|c|c|c|c|c|c|c}
\toprule 
\textbf{Systems} & \multicolumn{6}{c}{\textbf{RealNews}}    & \textbf{UDS-T}   \\
\cmidrule(lr){1-1} \cmidrule(lr){2-7} \cmidrule(lr){8-8}
        &         &         & \multicolumn{4}{c}{Typical Time}      &         \\
\cmidrule(lr){4-7}
        & Duration& Freq    & Day     & Week    & Month   & Season  & Duration\\
\cmidrule(lr){2-2}\cmidrule(lr){3-3}\cmidrule(lr){4-4}\cmidrule(lr){5-5}\cmidrule(lr){6-6}\cmidrule(lr){7-7}\cmidrule(lr){8-8}

BERT    & 1.33    & 1.68    & 1.75    & 1.53    & 3.78    & 0.87    & 1.77    \\
BERT + naive finetune & 1.21    & 1.45    & 1.47    & 1.28    & 3.28    & 1.08    & 2.06    \\
\modelname\ (ours)    & \textbf{0.75}        & \textbf{1.17}        & \textbf{1.72}        & \textbf{1.19}        & \textbf{3.42}        & \textbf{0.63}        & \textbf{1.49}        \\
\modelname\ (ours), normalized & 8\% & 13\% & 22\% & 17\% & 29\% & 16\% & 17\% \\
\toprule
\modelname\ -\keywordText{ADJ}    & 0.84          & 1.20          & 1.82          & 1.08          & 2.47          & 0.74          & 1.68          \\
\modelname\ -\keywordText{SL}     & 0.77          & 1.30          & 1.88          & 1.06          & 2.50          & 0.74          & 1.50          \\
\modelname\ -\keywordText{AUX}    & 0.77          & 1.28          & 1.61          & 1.31          & 3.25          & 0.78          & 1.51          \\
\modelname\ -\keywordText{MS}     & 0.84          & 1.12          & 1.82          & 1.5          & 3.17          & 0.61          & 1.69          \\
\modelname\ -\keywordText{AM} & 0.68          & 1.20          & 1.86          & 1.31          & 3.11          & 0.70          & 1.58  \\
\bottomrule 
\end{tabular}
\end{adjustbox}
\caption{Performance on intrinsic evaluations. 
% \danielk{I changed the name of the ``normalized row''}
The ``normalized'' row is the ratio of the distance to the gold label over the total number of labels in each dimension. 
Smaller is better. 
}
\label{tab:performance-intrinsic}
\end{table*}

The results on the filtered RealNews dataset and filtered UDST dataset are shown in Table~\ref{tab:performance-intrinsic}. We see that our proposed final model is mostly better than other variants, and achieves 19\% improvement over BERT on average on the normalized scale. 
%Table~\ref{tab:performance-intrinsic} also shows the results on the filtered UDST dataset, and we see that our proposed model outperforms BERT and all its variances. 

% To better understand the reason of improvements, we sampled 23 events in the RealNews dataset with a temporal argument indicating a duration of ``days''. We average the probabilities of BERT and our model's prediction on each possible labels, and we see a bell-shaped distribution in our model, while a discontinuous distribution from BERT. Intuitively, our model's distribution makes more sense as the distribution is sampled from multiple events. This is because our use of soft objective and label weight adjustments. 

We plot the embedding space of events with duration of ``seconds'' ``weeks'' or ``centuries'' in Fig~\ref{fig:bert-embedding-visualization} and Fig~\ref{fig:ours-embedding-visualization}. We take the verb's contextual representation, apply PCA to reduce the dimension from 768 to 50, and then t-SNE to reduce it further to 2. Comparing the two plots, we see that the clusters formed by BERT embeddings have a wider distribution over the space, and the clusters have more points in overlap, even though the three sets of events have drastically different duration values. Our proposed model's embedding is able to better cluster the events based on this temporal feature, which is expected.

\subsection{TimeBank Evaluation}
\label{subsec:timebank-eval}
Beyond unsupervised intrinsic experiments, we also evaluate the capability of the event temporal representation as a product of our model. That is, we finetune both BERT baseline and our model with the same process to compare the internal representations of the transformers. We use TimeML~\cite{saurei2005timeml,PanMuHo06}, a dataset with event duration annotated as lower and upper bounds. The task is to decide whether a given event has a duration longer or shorter than a day. This is a suitable task to evaluate the embeddings because deciding longer/shorter than a day requires reasoning with more than one label, and would also benefit from auxiliary dimensions like duration upper-bound.

The dataset contains 2,251 events, and we split the events based on sentences into 1,248/1,003 train/test. We formulate the training as a sequence classification task by taking the entire sentence and adding a special marker to the left of the verb indicating its position. The marker is unseen to both BERT and our model. We use the transformer output of the first token and feed it to an MLP for classification. We use a learning rate of 5e-5 and train for 3 epochs, and we repeat every reported number with 3 different random initialization and take the average.

\begin{table}[h]
\small
\centering
    \begin{tabu}{l|c|c|c}
        \toprule 
        System F1                                     & Accuracy & \textless{}Day F1 & $\geq$Day F1 \\ 
        \cmidrule(lr){1-1}                              \cmidrule(lr){2-2} \cmidrule(lr){3-3} \cmidrule(lr){4-4}
        BERT & 73.7 & 63.7 & 79.0  \\ 
        
        \modelname & \textbf{81.7} & \textbf{74.8} & \textbf{85.6} \\
        % SOTA (80/20 split) & - & 76 & 87 \\
        \bottomrule 
    \end{tabu}
\caption{Performance on TimeBank Classification}
\label{tab:performance-timebank}
\end{table}

Table~\ref{tab:performance-timebank} shows the results of the TimeBank experiment. We see around 7-11\% improvement over BERT on this task. Comparing with the state-of-the-art \cite{VempalaBlPa18} with a different training/testing split, our model is within 1.5\% of the best results but uses 25\% less training data.

\subsection{Subevent Relation Extraction}
\label{subsec:subevent_relations}
We apply our event representations to the task of event sub-super relation extraction. This is a proper evaluation because the task naturally benefits from temporal commonsense knowledge. Intuitively, short duration or high frequency indicates the event being at a lower hierarchy and vice versa. We test if the temporal focused event representations will improve.

We use HiEVE~\cite{glavas-etal-2014-hieve-a}, a dataset with annotations of four event relationships: no relation (NoRel), coreference (Coref), Child-Parent (C-P) and Parent-Child (P-C). There is no official split for this dataset, so we randomly 80/20 split the data at the document level and down-sample negative NoRel instances with a probability of 0.4.
%and completes the referential closure. That is, we complete some missing annotations of cases like event A coreferences with event B, event B as a sub-event of event C, yet no annotation on event A being a sub-event of event C. In our case, every C-P relation's reversed order is also guaranteed to be annotated P-C. We down-sample the negative NoRel instances with a probability of 0.4. 

Similarly, we formulate the problem as a sequence classification task, where two events are put into one sequence separated by ``\keywordCode{[SEP]},'' and verbs are marked by adding a marker token to their left. We use the representations of the first token and feed it to an MLP for classification. We train each model with a 5e-5 learning rate and 3 epochs. Each reported number is an average from 3 runs under different random initialization. During inference time, the probability scores for non-negative relations are averaged from the same event pair's sequences in both orders.

\bzch{Table~\ref{tab:performance-hieve} shows the results of the HiEVE experiment. We see that \modelname{} improves by 4\% and 8\% on the coreference task and the parent-child tasks over BERT, respectively.}

\begin{table}[h]
\small
\centering
    \begin{tabu}{lcccc}
        \toprule 
        Systems F1                                      & NoRel & Coref & C-P & P-C \\ 
        \cmidrule(lr){1-1}                              \cmidrule(lr){2-2}  \cmidrule(lr){3-3} \cmidrule(lr){4-4} \cmidrule(lr){5-5}
        BERT             & 90.5 & 47.9 & 40.7 & 40.6  \\ 
        \ignore{
        \rowfont{\color{red}}TmpSeq SS          &   90.6 & 47.5 & 45.7 & 44.4  \\ 
        \rowfont{\color{red}}TmpSeq SS AM   & 90.3 & 47.6 & 44.1 & 43.5 \\
        \rowfont{\color{red}}TmpSeq MS AM -SL   &  91.3 & 51.9 & 47.9 & 46.2 \\
        \rowfont{\color{red}}TmpSeq MS AM -ADJ & 91.1 & 50.9 & 50.2 & 49.1 \\
        }

        \modelname & \textbf{91.3} & \textbf{51.5} & \textbf{49.4} & \textbf{48.5} \\
        \bottomrule 
    \end{tabu}
\caption{Performance on HiEVE. The numbers are in percentages. Higher is better. }
\label{tab:performance-hieve}
\end{table}

\subsection{Temporal Question Answering}
We also evaluate on \mctaco{} \cite{ZKNR19}, a question answering dataset that requires comprehensive understandings of temporal common sense and reasoning. We compare the exact-match score across the 5 dimensions defined in \mctaco{}, although this work only focuses on 3 of them. We use the original baseline system and interchange transformer weights to compare between BERT and ours. However, because our model replaces temporal expressions with special tokens, it is at disadvantage to be directly evaluated on the original dataset with temporal expressions in natural language. To fix this, we run the same extraction system in \S\ref{subsec:cheap:supervision} with modifications to identify the dimension a question is asking, and augment candidate answers with our special tokens representing the temporal values (if any) mentioned. 
% \sout{distinguish duration and frequency questions}. 
This introduces rule-based dimension identification as well as coarse unit normalization to the systems, so we train/evaluated BERT baseline with the same modified data as well. Each number is an average of 5 runs with different random initializations.

\begin{table}[h]
\small
\centering
\begin{adjustbox}{width=0.48\textwidth}
    \begin{tabu}{l|c|c|c|c|c}
        \toprule 
        System & Duration & Ordering & Stationarity & Frequency & Typical Time \\ 
        \cmidrule(lr){1-1}                              \cmidrule(lr){2-2}  \cmidrule(lr){3-3} \cmidrule(lr){4-4} \cmidrule(lr){5-5} \cmidrule(lr){6-6}
        BERT & 33.4 & \textbf{36.5} & 57.6 & 43.3 & 39.5  \\ 
        \modelname & \textbf{34.6} & 35.1 & \textbf{57.9} & \textbf{45.1} & \textbf{40.9} \\
        \bottomrule 
    \end{tabu}
\end{adjustbox}
\caption{Performance on \mctaco{}. Numbers are percentages and indicate exact match (EM) metric. Higher is better. }
\label{tab:performance-mctaco}
\end{table}

Results on \mctaco{} are shown in Table~\ref{tab:performance-mctaco}. 
As expected, we find that our model achieves better performance on the three dimensions that are focused in this work (i.e., duration, frequency, and typical time) as well as stationarity. \bzch{However, the improvements are not very substantial, indicating the difficulty of this task and motivates future works.} The model \bzch{also} does slightly worse on ordering, which is worth investigating in future works. 

\section{Conclusion}
Temporal common sense (TCS) is an important yet challenging research topic. Despite the existence of several prior work on event duration, this is the first attempt to jointly model three key dimensions of TCS---duration, frequency, and typical time---
% \qn{\sout{simultaneously} a bit redundant given ``jointly''}\danielk{there is also another ``jointly'' in this sentence.} 
from cheap supervision signals mined from unannotated free text.
% freely-available \qn{I guess ``unannotated'' is more accurate} text. 
The proposed sequence modeling framework improves over BERT in terms of handling reporting bias, taking into account the ordinal relations and exploiting interactions among multiple dimensions of time. 
\bzch{The success of this model is confirmed by intrinsic evaluations on RealNews and UDS-T (where we see a 19\% improvement), as well as extrinsic evaluations on TimeBank, HiEVE and \mctaco. The proposed method may be an important module for future applications related to {\em time}.}
% \bz{highlight results}

% We propose a sequence pretraining method with novel techniques to improve language model's temporal representation jointly over several temporal dimensions. We show improvements on both intrinsic and extrinsic tasks.

% We also analyze the task of temporal common sense, which hopefully motivates future works to solve similar issues in a more general scope. Many dimensions, such as physical properties, share the same issues, and it's tempting to research on how we can build models that reasons better universally.

% \danielk{Mention something about potential extensions: for example, time-aware language modelling: having temporal objectives could result in better general language models. }
% \danielk{in some citations, conference names are only acronyms and sometimes its the extended conference name. Unify them? }
% \qn{good point, but this'd be very time consuming. it took me hours to do so for one earlier paper. i'm fine with the existence of imperfections in the bib}

\section*{Acknowledgments}
This research is based upon work supported in part by the office of the Director of National Intelligence (ODNI), Intelligence Advanced Research Projects Activity (IARPA), via IARPA Contract No. 2019-19051600006 under the BETTER Program and by Contract FA8750-19-2-1004 with the US Defense Advanced Research Projects Agency (DARPA). The views expressed are those of the authors and do not reflect the official policy or position of the Department of Defense or the U.S. Government.
This research is also supported by a grant from the Allen Institute for Artificial Intelligence (allenai.org).

\bibliography{acl2020}

\begin{thebibliography}{45}
\expandafter\ifx\csname natexlab\endcsname\relax\def\natexlab#1{#1}\fi

\bibitem[{Angeli et~al.(2012)Angeli, Manning, and Jurafsky}]{angeli2012parsing}
Gabor Angeli, Christopher~D Manning, and Daniel Jurafsky. 2012.
\newblock Parsing time: Learning to interpret time expressions.
\newblock In \emph{NAACL-HLT}, pages 446--455.

\bibitem[{Bagherinezhad et~al.(2016)Bagherinezhad, Hajishirzi, Choi, and
  Farhadi}]{bagherinezhad2016elephants}
Hessam Bagherinezhad, Hannaneh Hajishirzi, Yejin Choi, and Ali Farhadi. 2016.
\newblock Are elephants bigger than butterflies? reasoning about sizes of
  objects.
\newblock In \emph{AAAI}.

\bibitem[{Bauer et~al.(2018)Bauer, Wang, and Bansal}]{BauerWaBa18}
Lisa Bauer, Yicheng Wang, and Mohit Bansal. 2018.
\newblock Commonsense for generative multi-hop question answering tasks.
\newblock In \emph{EMNLP}, pages 4220--4230.

\bibitem[{Bethard et~al.(2016)Bethard, Savova, Chen, Derczynski, Pustejovsky,
  and Verhagen}]{BSCDPV16}
Steven Bethard, Guergana Savova, Wei-Te Chen, Leon Derczynski, James
  Pustejovsky, and Marc Verhagen. 2016.
\newblock {SemEval}-2016 {Task} 12: {Clinical TempEval}.
\newblock In \emph{SemEval}, pages 1052--1062.

\bibitem[{Chambers et~al.(2014)Chambers, Cassidy, McDowell, and
  Bethard}]{CCMB14}
Nathanael Chambers, Taylor Cassidy, Bill McDowell, and Steven Bethard. 2014.
\newblock Dense event ordering with a multi-pass architecture.
\newblock \emph{TACL}, 2:273--284.

\bibitem[{Chklovski and Pantel(2004)}]{ChklovskiPa04}
Timothy Chklovski and Patrick Pantel. 2004.
\newblock {VerbOcean: Mining the Web for Fine-Grained Semantic Verb Relations}.
\newblock In \emph{EMNLP}, pages 33--40.

\bibitem[{Cocos et~al.(2018)Cocos, Wharton, Pavlick, Apidianaki, and
  Callison-Burch}]{CWPAC18}
Anne Cocos, Veronica Wharton, Ellie Pavlick, Marianna Apidianaki, and Chris
  Callison-Burch. 2018.
\newblock Learning scalar adjective intensity from paraphrases.
\newblock In \emph{EMNLP}, pages 1752--1762.

\bibitem[{Devlin et~al.(2019)Devlin, Chang, Lee, and
  Toutanova}]{devlin2018bert}
Jacob Devlin, Ming-Wei Chang, Kenton Lee, and Kristina Toutanova. 2019.
\newblock {BERT}: Pre-training of deep bidirectional transformers for language
  understanding.
\newblock In \emph{NAACL}.

\bibitem[{Elazar et~al.(2019)Elazar, Mahabal, Ramachandran, Bedrax-Weiss, and
  Roth}]{elazar2019large}
Yanai Elazar, Abhijit Mahabal, Deepak Ramachandran, Tania Bedrax-Weiss, and Dan
  Roth. 2019.
\newblock How large are lions? inducing distributions over quantitative
  attributes.
\newblock In \emph{ACL}.

\bibitem[{Forbes and Choi(2017)}]{forbes2017verb}
Maxwell Forbes and Yejin Choi. 2017.
\newblock Verb physics: Relative physical knowledge of actions and objects.
\newblock In \emph{ACL}, volume~1, pages 266--276.

\bibitem[{Gardner et~al.(2018)Gardner, Grus, Neumann, Tafjord, Dasigi, Liu,
  Peters, Schmitz, and Zettlemoyer}]{gardner2018allennlp}
Matt Gardner, Joel Grus, Mark Neumann, Oyvind Tafjord, Pradeep Dasigi, Nelson~F
  Liu, Matthew Peters, Michael Schmitz, and Luke Zettlemoyer. 2018.
\newblock Allennlp: A deep semantic natural language processing platform.
\newblock In \emph{NLP-OSS}, pages 1--6.

\bibitem[{Glava{\v{s}} et~al.(2014)Glava{\v{s}}, {\v{S}}najder, Moens, and
  Kordjamshidi}]{glavas-etal-2014-hieve-a}
Goran Glava{\v{s}}, Jan {\v{S}}najder, Marie-Francine Moens, and Parisa
  Kordjamshidi. 2014.
\newblock {H}i{E}ve: A corpus for extracting event hierarchies from news
  stories.
\newblock In \emph{LREC}, pages 3678--3683, Reykjavik, Iceland. European
  Language Resources Association (ELRA).

\bibitem[{Gordon and Van~Durme(2013)}]{GordenDu13}
Jonathan Gordon and Benjamin Van~Durme. 2013.
\newblock Reporting bias and knowledge acquisition.
\newblock In \emph{AKBC}, pages 25--30. ACM.

\bibitem[{Graff et~al.(2003)Graff, Kong, Chen, and Maeda}]{graff2003english}
David Graff, Junbo Kong, Ke~Chen, and Kazuaki Maeda. 2003.
\newblock English gigaword.
\newblock \emph{Linguistic Data Consortium, Philadelphia}, 4(1):34.

\bibitem[{Granroth-Wilding and Clark(2016)}]{GranrothCl16}
Mark Granroth-Wilding and Stephen~Christopher Clark. 2016.
\newblock What happens next? event prediction using a compositional neural
  network model.
\newblock In \emph{ACL}.

\bibitem[{Gusev et~al.(2011)Gusev, Chambers, Khaitan, Khilnani, Bethard, and
  Jurafsky}]{GCKKBJ11}
Andrey Gusev, Nathanael Chambers, Pranav Khaitan, Divye Khilnani, Steven
  Bethard, and Dan Jurafsky. 2011.
\newblock Using query patterns to learn the duration of events.
\newblock In \emph{IWCS}, pages 145--154.

\bibitem[{Huang et~al.(2019)Huang, Sun, Qiu, and Huang}]{Huang2019GlossBERTBF}
Luyao Huang, Chi Sun, Xipeng Qiu, and Xuanjing Huang. 2019.
\newblock {G}loss{BERT}: {BERT} for word sense disambiguation with gloss
  knowledge.
\newblock In \emph{EMNLP}, pages 3507--3512.

\bibitem[{Lee et~al.(2014)Lee, Artzi, Dodge, and Zettlemoyer}]{LADZ14}
Kenton Lee, Yoav Artzi, Jesse Dodge, and Luke Zettlemoyer. 2014.
\newblock Context-dependent semantic parsing for time expressions.
\newblock In \emph{ACL (1)}, pages 1437--1447.

\bibitem[{Leeuwenberg and Moens(2017)}]{LeeuwenbergMo17}
Artuur Leeuwenberg and Marie-Francine Moens. 2017.
\newblock Structured learning for temporal relation extraction from clinical
  records.
\newblock In \emph{EACL}.

\bibitem[{Leeuwenberg and Moens(2018)}]{LeeuwenbergMo18}
Artuur Leeuwenberg and Marie-Francine Moens. 2018.
\newblock Temporal information extraction by predicting relative time-lines.
\newblock \emph{EMNLP}.

\bibitem[{Li et~al.(2018)Li, Ding, and Liu}]{LiDiLi18}
Zhongyang Li, Xiao Ding, and Ting Liu. 2018.
\newblock Constructing narrative event evolutionary graph for script event
  prediction.
\newblock \emph{IJCAI}.

\bibitem[{Liu et~al.(2019)Liu, Ott, Goyal, Du, Joshi, Chen, Levy, Lewis,
  Zettlemoyer, and Stoyanov}]{liu2019roberta}
Yinhan Liu, Myle Ott, Naman Goyal, Jingfei Du, Mandar Joshi, Danqi Chen, Omer
  Levy, Mike Lewis, Luke Zettlemoyer, and Veselin Stoyanov. 2019.
\newblock Roberta: A robustly optimized bert pretraining approach.
\newblock \emph{arXiv preprint arXiv:1907.11692}.

\bibitem[{Llorens et~al.(2015)Llorens, Chambers, UzZaman, Mostafazadeh, Allen,
  and Pustejovsky}]{LCUMAP15}
Hector Llorens, Nathanael Chambers, Naushad UzZaman, Nasrin Mostafazadeh, James
  Allen, and James Pustejovsky. 2015.
\newblock {SemEval}-2015 {Task} 5: {QA} {TEMPEVAL} - evaluating temporal
  information understanding with question answering.
\newblock In \emph{SemEval}, pages 792--800.

\bibitem[{Ning et~al.(2017)Ning, Feng, and Roth}]{NingFeRo17}
Qiang Ning, Zhili Feng, and Dan Roth. 2017.
\newblock A structured learning approach to temporal relation extraction.
\newblock In \emph{EMNLP}.

\bibitem[{Ning et~al.(2018{\natexlab{a}})Ning, Feng, Wu, and Roth}]{NFWR18}
Qiang Ning, Zhili Feng, Hao Wu, and Dan Roth. 2018{\natexlab{a}}.
\newblock Joint reasoning for temporal and causal relations.
\newblock In \emph{ACL}.

\bibitem[{Ning et~al.(2018{\natexlab{b}})Ning, Wu, Peng, and Roth}]{NWPR18}
Qiang Ning, Hao Wu, Haoruo Peng, and Dan Roth. 2018{\natexlab{b}}.
\newblock Improving temporal relation extraction with a globally acquired
  statistical resource.
\newblock In \emph{NAACL}, pages 841--851.

\bibitem[{Ning et~al.(2018{\natexlab{c}})Ning, Zhou, Feng, Peng, and
  Roth}]{NZFPR18}
Qiang Ning, Ben Zhou, Zhili Feng, Haoruo Peng, and Dan Roth.
  2018{\natexlab{c}}.
\newblock {CogCompTime}: A tool for understanding time in natural language.
\newblock In \emph{EMNLP}.

\bibitem[{Pan et~al.(2006)Pan, Mulkar, and Hobbs}]{PanMuHo06}
Feng Pan, Rutu Mulkar, and Jerry~R Hobbs. 2006.
\newblock Extending {TimeML} with typical durations of events.
\newblock In \emph{ARTE}, pages 38--45. Association for Computational
  Linguistics.

\bibitem[{Peng et~al.(2019)Peng, Ning, and Roth}]{PengNiRo19}
Haoruo Peng, Qiang Ning, and Dan Roth. 2019.
\newblock {KnowSemLM: A Knowledge Infused Semantic Language Model}.
\newblock In \emph{CoNLL}.

\bibitem[{Peters et~al.(2018)Peters, Neumann, Iyyer, Gardner, Clark, Lee, and
  Zettlemoyer}]{peters2018deep}
Matthew~E Peters, Mark Neumann, Mohit Iyyer, Matt Gardner, Christopher Clark,
  Kenton Lee, and Luke Zettlemoyer. 2018.
\newblock Deep contextualized word representations.
\newblock In \emph{Proceedings of NAACL-HLT}, pages 2227--2237.

\bibitem[{Pustejovsky et~al.(2003)Pustejovsky, Hanks, Sauri, See, Gaizauskas,
  Setzer, Radev, Sundheim, Day, Ferro et~al.}]{PHSSGSRSDFo03}
James Pustejovsky, Patrick Hanks, Roser Sauri, Andrew See, Robert Gaizauskas,
  Andrea Setzer, Dragomir Radev, Beth Sundheim, David Day, Lisa Ferro, et~al.
  2003.
\newblock The {TIMEBANK} corpus.
\newblock In \emph{Corpus linguistics}, volume 2003, page~40.

\bibitem[{Saur{\'e}i et~al.(2005)Saur{\'e}i, Littman, Knippen, Gaizauskas,
  Setzer, and Pustejovsky}]{saurei2005timeml}
Roser Saur{\'e}i, Jessica Littman, Bob Knippen, Robert Gaizauskas, Andrea
  Setzer, and James Pustejovsky. 2005.
\newblock Timeml annotation guidelines.

\bibitem[{Schubert(2002)}]{Schubert02}
Lenhart Schubert. 2002.
\newblock Can we derive general world knowledge from texts?
\newblock In \emph{HLT}, pages 94--97. Morgan Kaufmann Publishers Inc.

\bibitem[{Shi and Lin(2019)}]{shi2019simple}
Peng Shi and Jimmy Lin. 2019.
\newblock Simple bert models for relation extraction and semantic role
  labeling.
\newblock \emph{arXiv preprint arXiv:1904.05255}.

\bibitem[{Str{\"o}tgen and Gertz(2010)}]{strotgen2010heideltime}
Jannik Str{\"o}tgen and Michael Gertz. 2010.
\newblock {HeidelTime}: High quality rule-based extraction and normalization of
  temporal expressions.
\newblock In \emph{SemEval}, pages 321--324. Association for Computational
  Linguistics.

\bibitem[{Tandon et~al.(2018)Tandon, Dalvi, Grus, Yih, Bosselut, and
  Clark}]{TDGYBC18}
Niket Tandon, Bhavana Dalvi, Joel Grus, Wen-tau Yih, Antoine Bosselut, and
  Peter Clark. 2018.
\newblock Reasoning about actions and state changes by injecting commonsense
  knowledge.
\newblock In \emph{EMNLP}, pages 57--66.

\bibitem[{UzZaman et~al.(2013)UzZaman, Llorens, Allen, Derczynski, Verhagen,
  and Pustejovsky}]{ULADVP13}
Naushad UzZaman, Hector Llorens, James Allen, Leon Derczynski, Marc Verhagen,
  and James Pustejovsky. 2013.
\newblock {SemEval}-2013 {Task} 1: {TEMPEVAL}-3: Evaluating time expressions,
  events, and temporal relations.
\newblock \emph{*SEM}, 2:1--9.

\bibitem[{Van~Durme(2009)}]{Durme09}
Benjamin Van~Durme. 2009.
\newblock \emph{Extracting implicit knowledge from text}.
\newblock Ph.D. thesis, University of Rochester.

\bibitem[{Vashishtha et~al.(2019)Vashishtha, Van~Durme, and
  White}]{VashishthaVaWh19}
Siddharth Vashishtha, Benjamin Van~Durme, and Aaron~Steven White. 2019.
\newblock Fine-grained temporal relation extraction.
\newblock In \emph{ACL}, pages 2906--2919.

\bibitem[{Vempala et~al.(2018)Vempala, Blanco, and Palmer}]{VempalaBlPa18}
Alakananda Vempala, Eduardo Blanco, and Alexis Palmer. 2018.
\newblock Determining event durations: Models and error analysis.
\newblock In \emph{NAACL}, volume~2, pages 164--168.

\bibitem[{Williams(2012)}]{williams2012extracting}
Jennifer Williams. 2012.
\newblock Extracting fine-grained durations for verbs from twitter.
\newblock In \emph{ACL-SRW}, pages 49--54. Association for Computational
  Linguistics.

\bibitem[{Yang et~al.(2018)Yang, Birnbaum, Wang, and
  Downey}]{yang2018extracting}
Yiben Yang, Larry Birnbaum, Ji-Ping Wang, and Doug Downey. 2018.
\newblock Extracting commonsense properties from embeddings with limited human
  guidance.
\newblock In \emph{ACL}, volume~2, pages 644--649.

\bibitem[{Zellers et~al.(2019)Zellers, Holtzman, Rashkin, Bisk, Farhadi,
  Roesner, and Choi}]{zellers2019neuralfakenews}
Rowan Zellers, Ari Holtzman, Hannah Rashkin, Yonatan Bisk, Ali Farhadi,
  Franziska Roesner, and Yejin Choi. 2019.
\newblock Defending against neural fake news.
\newblock In \emph{NeurIPS}, pages 9054--9065. Curran Associates, Inc.

\bibitem[{Zhang et~al.(2017)Zhang, Rudinger, Duh, and Van~Durme}]{ZRDV17}
Sheng Zhang, Rachel Rudinger, Kevin Duh, and Benjamin Van~Durme. 2017.
\newblock Ordinal common-sense inference.
\newblock \emph{TACL}, 5(1):379--395.

\bibitem[{Zhou et~al.(2019)Zhou, Khashabi, Ning, and Roth}]{ZKNR19}
Ben Zhou, Daniel Khashabi, Qiang Ning, and Dan Roth. 2019.
\newblock {"Going on a vacation" takes longer than "Going for a walk": A Study
  of Temporal Commonsense Understanding}.
\newblock In \emph{EMNLP}.

\end{thebibliography}
\bibliographystyle{acl_natbib}

% \section{Appendices}

\end{document}